\pdfoutput=1
\documentclass[11pt]{article}
\usepackage{arabtex}

\usepackage{acl}

\usepackage{times}
\usepackage{latexsym}
\usepackage[linesnumbered,ruled,lined]{algorithm2e}
\usepackage{amsmath,amssymb,amsfonts}
\usepackage{graphicx}
\usepackage{textcomp}
\usepackage{subcaption}
\usepackage{multirow}
\usepackage{rotating}
\usepackage{colortbl}
\usepackage{tikz}
\usepackage{soul}
\usepackage{bbm}
\usepackage{booktabs} % To thicken table lines
\usepackage{adjustbox}
\usepackage{comment}
\usepackage{arydshln}
\usepackage{tcolorbox}
\usepackage{utf8}
\usepackage{enumerate} 
\hypersetup{breaklinks=true}

\usepackage[T1]{fontenc}
\usepackage[utf8]{inputenc}

\usepackage{microtype}

\usepackage{inconsolata}

\title{On The Origin of Cultural Biases in Language Models: \\ From Pre-training Data to Linguistic Phenomena}

\author{Tarek Naous \and Wei Xu \\
  Georgia Institute of Technology \\
 \texttt{tareknaous@gatech.edu; wei.xu@cc.gatech.edu}
 }

\begin{document}
\maketitle

\begin{abstract}
Language Models (LMs) have been shown to exhibit a strong preference towards entities associated with Western culture when operating in non-Western languages.  In this paper, we aim to uncover the origins of entity-related cultural biases in LMs by analyzing several contributing factors, including the representation of entities in pre-training data and the impact of variations in linguistic phenomena across languages. We introduce CAMeL-2, a parallel Arabic-English benchmark of 58,086 entities associated with Arab and Western cultures and 367 masked natural contexts for entities. Our evaluations using CAMeL-2 reveal reduced performance gaps between cultures by LMs when tested in English compared to Arabic. We find that LMs struggle in Arabic with entities that appear at high frequencies in pre-training, where entities can hold multiple word senses. This also extends to entities that exhibit high lexical overlap with languages that are not Arabic but use the Arabic script. Further, we show how frequency-based tokenization leads to this issue in LMs, which gets worse with larger Arabic vocabularies. We will make CAMeL-2 available at: \url{https://github.com/tareknaous/camel2}. 
\end{abstract}

% \url{https://anonymized_url.org}

\section{Introduction}

Multilingual Language Models (LMs) are playing a crucial role in making AI technology accessible to global communities \cite{ustun2024aya, singh-etal-2024-aya}. As these communities represent diverse cultural backgrounds, the multilingual challenge for LMs does not merely stop at handling different languages \cite{blevins2024breakingcursemultilingualitycrosslingual}, but extends to capturing cultural nuances \cite{adilazuarda2024towards}. However, past research has highlighted strong favoritism in LMs towards entities associated with Western culture when operating in non-Western languages, leading to a struggle by LMs to adapt to cultural contexts and gaps in their performance between cultures on NLP tasks \cite{naous-etal-2024-beer}.

\begin{figure}[t]
    \centering
    \includegraphics[width=\linewidth]{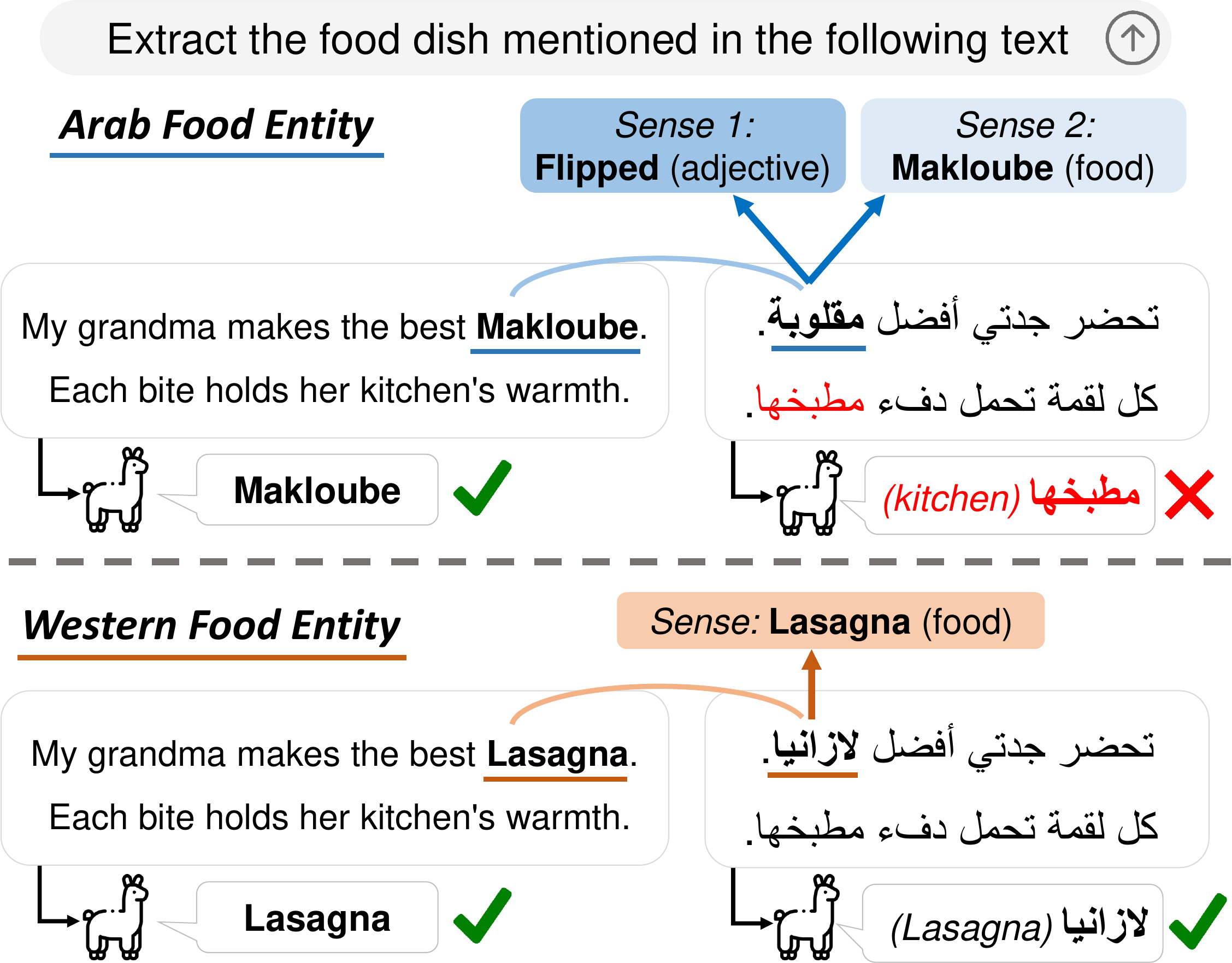}
    \caption{Responses of a LM (\includegraphics[height=1em]{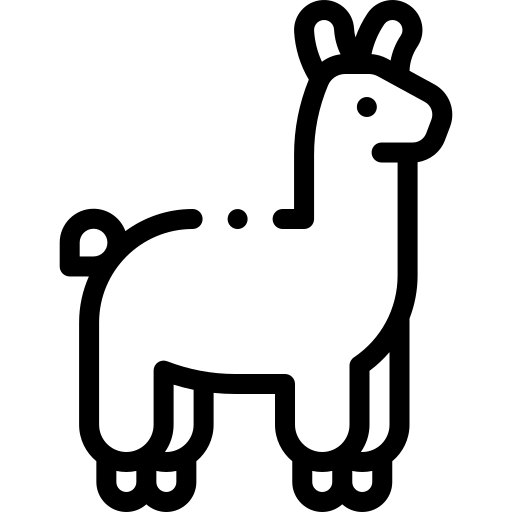}) tasked to extract the food dish from the same text in English and Arabic. The LM identifies the Arab dish \textit{``Makloube''} in English, but fails in Arabic where the word \textit{``Makloube''} holds two senses. The LM does not struggle with the Western dish \textit{``Lasagna''} which holds only one sense in both languages.}
    \label{fig:figure-intro}
    \vspace{-.2cm}
\end{figure}

While entity-related biases in LMs have been traditionally studied as a reflection of imbalanced representations in pre-training data \cite{gallegos2024bias, li2024attributing}, it is often overlooked how linguistic phenomena in non-English languages can also incite those biases.  For example, when an LM is asked to extract an Arab food dish such as \textit{“Makloube”} from text, it can fail to do so in Arabic where the word used for the dish holds two senses (as the food dish or as the adjective \textit{``flipped”}), but can successfully extract the same dish from the parallel English text, as shown in Figure~\ref{fig:figure-intro}. Yet, within the same context, the LM does not face this struggle when we replace \textit{“Makloube”} with a Western dish \textit{“Lasagna”}, which holds only one sense of a food dish both in English and in Arabic. These observations raise the question: \textbf{do varying linguistic phenomena exhibited by cultural entities influence cultural biases in LMs?}

% Our study aims to uncover the origins of entity-related biases in LMs by analyzing how cross-linguistic differences impact their performance on entities associated with different cultures.

% We explore a variety of factors that may lead to this disparity between both languages (\S\ref{sec:origin-of-biases}). Our analyses show that LMs struggle at recognizing entities associated with Arab culture which appear at very high frequencies in Arabic pre-training data (\S\ref{subsec:corpus-frequencies}). We find that such high-frequency entities exhibit strong word polysemy in Arabic, posing a challenge for LMs in recognizing entities and adapting to cultural contexts  (\S\ref{subsec:entity-polysemy-analysis}).

\begin{table*}[t!]
\centering
\setcode{utf-8}
\resizebox{0.9\linewidth}{!}{
\begin{tabular}{@{}clc@{}}
\toprule
\textbf{Food Contexts} &  & \textbf{Location Contexts} \\ \midrule
\multicolumn{3}{c}{\textbf{Text-Infilling \& NER} - CAMeL \cite{naous-etal-2024-beer}} \\ \hdashline[1pt/1pt]
\small{\<كتير طيبة>} \texttt{[MASK]} \small{\<الغدا عربي اليوم عملت>} &  & \small{\<العربية و هي في غاية الروعة>} \texttt{[MASK]} \small{\<انا منذ ايام كنت في مدينة>} \\
\small{(Today's lunch is Arab, I've cooked \texttt{[MASK]} which is very delicious)} &  & \small{(I was in the Arab city of \texttt{[MASK]} a few days ago and it is incredibly wonderful)} \\ \midrule
\multicolumn{3}{c}{\textbf{Extractive QA} - CAMeL-2 (this work)} \\ \hdashline[1pt/1pt]
\begin{tabular}[c]{@{}c@{}}\small{\<بإشراف امي لانها معانده إلا وتطبخ  بنفسها>} \texttt{[MASK]} \small{\<قررت ادخل اسوي>}\\ \small{\<لان ولدها المفضل بخاطره مالح وهي تعبانه وانا ماريدها تتعب اكثر>}\end{tabular} &  & \begin{tabular}[c]{@{}c@{}}\small{\<حيث تم عرض مشاكل>} \texttt{[MASK]} \small{\<استقبل الشيخ بهاء مساء اليوم وفدا من أهالي>}\\ \small{\<وسبل معالجتها. من جهته جدد الشيخ تعهده بحل هذه المشاكل والبدء بنهضة جديدة>}\end{tabular} \\
\begin{tabular}[c]{@{}c@{}}\small{(I decided to go in and make \texttt{[MASK]} under the supervision of my mother,}\\ \small{because she insists to cook only by herself since her favorite son has salty} \\ \small{taste, and she is tired, and I don’t want her to get more tired)}\end{tabular} &  & \begin{tabular}[c]{@{}c@{}}\small{(Sheikh Bahaa received this evening a delegation of people from \texttt{[MASK]} where }\\ \small{problems and ways to address them in were presented. For his part, the Sheikh  }\\  \small{renewed his pledge to solving these problems and starting a new renaissance)}\end{tabular} \\ \bottomrule
\end{tabular}
}
\caption{Example food and location contexts collected in CAMeL-2 for Extractive QA evaluation, compared with contexts from CAMeL \cite{naous-etal-2024-beer}. Extractive QA contexts are longer and mention entities more implicitly.}
\label{tab:contexts-examples}
\end{table*}

Our study investigates the impact of such cross-linguistic differences to uncover the origins of entity-related biases in LMs. To enable our analyses, we introduce CAMeL-2, a parallel Arabic-English resource of 58,086 entities associated with Arab and Western cultures across seven entity types, and a set of 367 natural contexts for these entities (\S\ref{sec:camel2}). Using CAMeL-2, we evaluate a variety of LMs on extractive QA and NER, revealing smaller performance gaps between cultures when LMs are tested in English compared to Arabic (\S\ref{sec:bias-comparison-arabic-english}).

Our analyses show that LMs struggle at recognizing entities associated with Arab culture which appear at very high frequencies in Arabic pre-training data (\S\ref{subsec:corpus-frequencies}), where such entities exhibit strong word polysemy in Arabic (\S\ref{subsec:entity-polysemy-analysis}). We also find that high lexical overlap of Arab entities with pre-training data of languages that use Arabic script (e.g., Farsi, Urdu, etc.) causes drops in performance (\S\ref{subsec:lexical-overlap}). Lastly, we show how tokenization causes LMs to struggle with Arab entities when tokenized into a single token, an issue that worsens with larger Arabic vocabularies (\S\ref{subsec:tokenization-analysis}).

\section{CAMeL-2: A Parallel Benchmark}
\label{sec:camel2}

Our goal is to investigate whether LMs handle entities associated with Arab and Western cultures differently when tested in Arabic \textit{vs.} English languages (\S \ref{sec:bias-comparison-arabic-english}). To do this, we first construct a \textit{bilingual} version of the entity-centric CAMeL \cite{naous-etal-2024-beer} benchmark for measuring cultural biases in LMs and extend its coverage by three times (\S\ref{subsec:enhancing-coverage}). Table~\ref{tab:stats-camel2} compares the statistics of CAMeL-2 vs. CAMeL. We also collect longer contexts where these entities are mentioned for evaluating LMs in a more challenging setup of extractive QA (\S\ref{subsec:natural-contexts}).

\subsection{About CAMeL}

The original CAMeL benchmark \cite{naous-etal-2024-beer} consists of 20,249 cultural entities extracted from Wikidata and web-crawl data and annotated with Arab or Western cultural association. CAMeL also includes a set of textual contexts where these entities may naturally occur (see examples in Table \ref{tab:contexts-examples}), derived from Arabic X/Twitter data. All entities and contexts in CAMeL are written in the Arabic language \textit{only}, limiting the ability to test the behavior of LMs when being prompted in English. To enable such comparisons, we construct a parallel extension to CAMeL by not only adding the English translation of each entity and context, but also increasing the overall number of entities and length of contexts. We find that Wikipedia contains more entities relevant to Arab culture than Wikidata for \textit{authors}, \textit{beverage}, \textit{food}, \textit{names}, \textit{religious places}, and \textit{sports clubs} (\S\ref{subsec:enhancing-coverage}). We expand the coverage of \textit{location} entities using public geographic data.

\begin{table}[t]
\small
\centering
\begin{adjustbox}{width=0.85\linewidth}
\begin{tabular}{@{}lllr@{}}
\toprule
\textbf{Entity Type} & \multicolumn{1}{c}{\textbf{CAMeL}} & \multicolumn{1}{c}{\textbf{CAMeL-2}} & \multicolumn{1}{c}{\textbf{Increase}} \\ \midrule
Authors & 571 & 6,315 & 11.05\scalebox{0.7}{$\times$} \\
Beverage & 142 & 255 & 1.79\scalebox{0.7}{$\times$} \\
Food & 578 & 2,283 & 3.94\scalebox{0.7}{$\times$} \\
Locations & 12,497 & 35,200 & 2.81\scalebox{0.7}{$\times$} \\
Names & 1,533 & 3,842 & 2.50\scalebox{0.7}{$\times$} \\
Religious & 2,428 & 5,049 & 2.07\scalebox{0.7}{$\times$} \\
Sports Clubs & 2,500 & 5,142 & 2.05\scalebox{0.7}{$\times$} \\ \midrule
\textbf{Total} & 20,249 & 58,086 & 2.86\scalebox{0.7}{$\times$} \\ \bottomrule
\end{tabular}
\end{adjustbox}
\caption{Number of entities in the bilingual CAMeL-2 (this work) vs. the monolingual CAMeL \cite{naous-etal-2024-beer}. We increase the size of the benchmark by 2.86\scalebox{0.7}{$\times$}.}
\label{tab:stats-camel2}
\end{table}

\subsection{Collecting Cultural Entities}
\label{subsec:enhancing-coverage}

\paragraph{Entity Extraction from Wikipedia.}  We leverage the categorization feature in Wikipedia and identify, for each entity type, a generic category that is repeatedly used to group together articles relevant to a specific country. For example, the ``\texttt{[country adjective] cuisine}'' category encompasses all food-related articles of a country (e.g., \textit{Syrian} cuisine, \textit{Irish} cuisine, etc.). We extract the titles of all articles associated with each country. The entities are then obtained from the article titles, which in most cases are direct references to the entity of interest. Additionally, we extract the body of text for each article which we use for distantly supervised fine-tuning of NER models (\S\ref{subsec:fairness-task}). See Appendix~\ref{appendix:camel2} for details and the categories used.

\paragraph{Annotation of Cultural Entities.} The extracted entities from Wikipedia were then manually annotated for cultural association. We hired two college students who are native Arabic speakers to classify entities into:  \textit{Arab culture} (Arab countries), \textit{Western culture} (European and North American countries), \textit{other cultures}, or \textit{not culture-specific}. The annotator agreement is 0.825 by Cohen's Kappa. The cases of disagreement were discussed in an adjudication step to decide on the final label.

\begin{figure*}[t]
    \centering
    \includegraphics[width=0.95\linewidth]{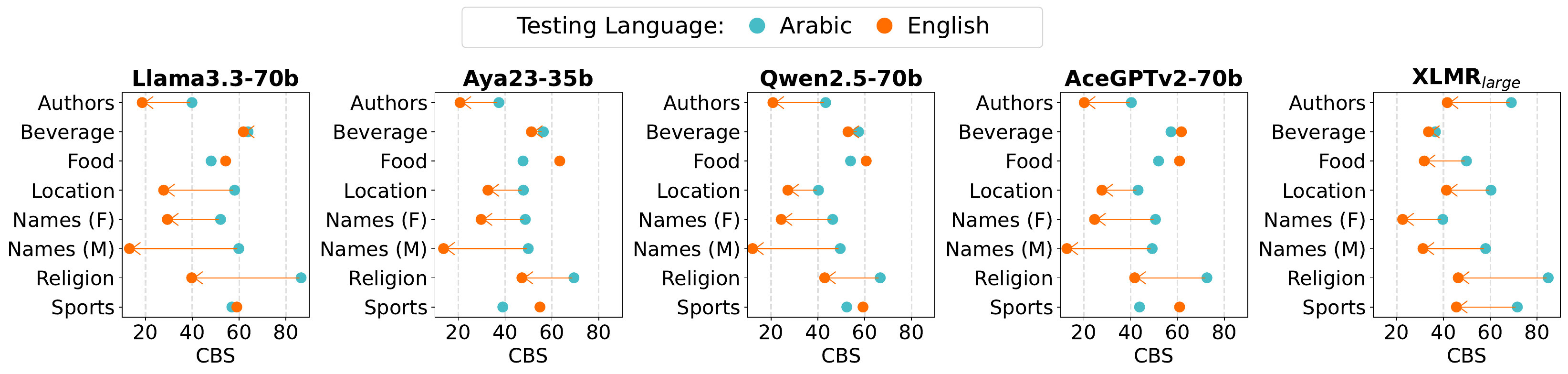}
    \caption{\textbf{C}ultural \textbf{B}ias \textbf{S}core ($\downarrow$) (\S \ref{subsec:text-infilling-task}) per entity type on culturally-grounded contexts from CAMeL-2. LMs can adapt better to Arab culture when tested in English.}
    \label{fig:cbs-comparison-arabic-english}
\end{figure*}

\paragraph{Mapping Arabic Entities to English.} Since both Wikidata and Wikipedia support multiple languages, we automatically map the Arabic entities to their English versions, which were available for 88.34\% of entities in CAMel-2. For the remainder of the entities which were only available in Arabic, we manually search for their commonly used English transliterations on the Internet. For example, a Tunisian sports club  \setcode{utf-8}``\<المظيلة>'' can be transliterated as ``\textit{Mzilla}'' or ``\textit{Mdhilla}'', both of which are valid. However, the form used by Tunisians is ``\textit{Mdhilla}'' as it aligns with their phonetic interpretation of Arabic letters in the Tunisian dialect. 

\paragraph{Georgraphic Data-based Extraction.} The Arab \textit{location} entities extracted from Wikidata in CAMeL had relatively low representation at 1,061 locations compared to 11,436 Western locations. We extract additional locations for all Arab countries from the OpenStreetMap\footnote{\url{https://www.openstreetmap.org/}} (OSM) database, which provides Arabic-English pairs of locations in Arab countries.\footnote{We determine Arab countries based on the league of Arab states: \url{https://arabmpi.org/en/home}}  This resulted in an extensive set of 23,765 Arab locations, enabling further analyses across Arab regions that were influenced by other languages in the history (\S \ref{subsec:entity-polysemy-analysis}).

 % \begin{tabular}{@{}lccc!{\vrule}ccc!{\vrule}ccc!{\vrule}ccc@{}}

\begin{table*}[]
\small
\centering
\resizebox{0.9\textwidth}{!}{%
\begin{tabular}{@{}lccc!{\vrule}ccc!{\vrule}ccc!{\vrule}ccc@{}}
\toprule
 & \multicolumn{6}{c}{\textbf{Llama3.3-70b}} & \multicolumn{6}{c}{\textbf{XLMR$_{\mathrm{large}}$}} \\ \cmidrule(l){2-13} 
 & \multicolumn{3}{c}{\textbf{Arabic}} & \multicolumn{3}{c}{\textbf{English}} & \multicolumn{3}{c}{\textbf{Arabic}} & \multicolumn{3}{c}{\textbf{English}} \\ \cmidrule(l){2-13} 
 & \textit{Arab} & \textit{Western} & $\Delta$Acc & \textit{Arab} & \textit{Western} & $\Delta$Acc & \textit{Arab} & \textit{Western} & $\Delta$F1 & \textit{Arab} & \textit{Western} & $\Delta$F1 \\ \midrule
Authors & 92.62 & 90.28 & -2.34 & 98.99 & 99.16 & \cellcolor[HTML]{FFF3EA}0.17 & 86.80 & \textbf{87.93} & \cellcolor[HTML]{DCF3F4}1.13 & 95.64 & 94.98 & -0.66 \\
Beverage & 82.65 & 78.19 & -4.46 & 99.14 & 97.71 & -1.43 & 63.06 & \textbf{72.86} & \cellcolor[HTML]{98DBE0}9.80 & 92.06 & 89.77 & -3.29 \\
Food & 84.08 & \textbf{84.71} & \cellcolor[HTML]{DCF3F4}0.63 & 95.84 & 98.21 & \cellcolor[HTML]{FFF3EA}2.37 & 63.76 & \textbf{73.59} & \cellcolor[HTML]{98DBE0}9.83 & 91.57 & 90.45 & -1.12 \\
Location & 80.66 & \textbf{95.59} & \cellcolor[HTML]{80D2D8}14.93 & 98.58 & 99.89 & \cellcolor[HTML]{FFF3EA}1.31 & 64.07 & \textbf{91.32} & \cellcolor[HTML]{46BDC6}27.25 & 89.54 & 95.71 & \cellcolor[HTML]{FFDDC4}6.17 \\
Names (F) & 63.38 & \textbf{77.39} & \cellcolor[HTML]{46BDC6}14.01 & 99.86 & 99.14 & -0.72 & 62.22 & \textbf{82.65} & \cellcolor[HTML]{46BDC6}20.43 & 97.87 & 96.36 & -1.51 \\
Names (M) & 75.45 & \textbf{76.23} & \cellcolor[HTML]{DCF3F4}0.78 & 99.43 & 99.78 & \cellcolor[HTML]{FFF3EA}0.35 & 80.09 & \textbf{85.03} & \cellcolor[HTML]{80D2D8}4.94 & 94.01 & 93.13 & -0.88 \\
Sports & 68.58 & \textbf{79.01} & \cellcolor[HTML]{46BDC6}10.43 & 92.77 & 96.02 & \cellcolor[HTML]{FFF3EA}3.25 & 74.52 & \textbf{84.14} & \cellcolor[HTML]{80D2D8}9.62 & 92.14 & 93.12 & \cellcolor[HTML]{FFF3EA}0.97 \\
Religious & 82.49 & \textbf{82.98} & \cellcolor[HTML]{DCF3F4}0.49 & 98.52 & 97.69 & -0.83 & 95.30 & \textbf{97.13} & \cellcolor[HTML]{DCF3F4}1.83 & 94.34 & 95.76 & \cellcolor[HTML]{FFF3EA}1.42 \\ \bottomrule
\end{tabular}
}
\caption{Average performance of Llama3.3-70b (QA Accuracy $\uparrow$) and XLMR$_{\mathrm{large}}$ (NER F1 $\uparrow$) on Arab and Western entities when tested in Arabic and English. More results with Aya23-35b and AceGPTv2-70b are in Appendix. $\Delta$Acc and $\Delta$F1 are performance differences between Western and Arab entities. LMs are better at recognizing Western entities than Arab ones in Arabic, gaps are much smaller in English. }
\label{tab:fairness-comparison-arabic-english}
%\vspace{-.1in}
\end{table*}

\subsection{Constructing Contexts for Extractive QA}
\label{subsec:natural-contexts}

CAMeL provides 250 masked contexts where only entities associated with Arab culture are appropriate \texttt{[MASK]} token fillings. This allows evaluation of LMs on adaptation to cultural contexts and on NER of entities with different cultural associations \cite{naous-etal-2024-beer}, but \textit{only} for Arabic language. Further, these contexts are short and explicitly refer to the masked entity, making them less suitable for evaluating GPT-type LMs on tasks such as extractive QA, which aligns better with their usage. To address this limitation, we collect 117 new, longer contexts from the X/Twitter platform where entities are mentioned more implicitly. This presents a challenging evaluation setup for LMs where understanding of context is necessary for extracting the entity (see comparative examples in Table~\ref{tab:contexts-examples}).

For each entity type, we perform keyword search using 30 randomly sampled Arabic entities to capture natural discussions about entities. We search over the two months period of 7/1/2024 to 9/1/2024 and manually inspect tweets to identify ones that are long and make an indirect reference to the entity. From these, we construct 10 to 20 extractive QA contexts for each entity type by replacing the user-mentioned entities by a \texttt{[MASK]}. To enable comparative Arabic-English evaluations, we manually translate each culturally-grounded context from the original CAMeL and the newly collected QA contexts from Arabic to English.

\section{Is Western Bias in LMs Consistent Across Arabic and English Languages?}
\label{sec:bias-comparison-arabic-english}

We start by studying whether LMs show the same degree of favoritism for Western culture, when operating in Arabic vs. English languages. Our analyses focus on two aspects: cultural adaptation in text-infilling (\S\ref{subsec:text-infilling-task}) and cross-cultural performance on extractive QA and NER (\S\ref{subsec:fairness-task}).

\subsection{Language Models}
\label{subsec:language-models}

We experiment with LMs that have been trained on both Arabic and English. We use the recent multilingual LMs of \textbf{Llama-3.3} \cite{dubey2024llama},  \textbf{Aya-23} \cite{aryabumi2024aya}, \textbf{Qwen-2.5} \cite{yang2024qwen2}, as well as \textbf{AceGPTv1.5} \cite{zhu2024second} which expands the Arabic vocabulary of Llama-2 and further fine-tunes it on Arabic data, \textbf{AceGPTv2} \cite{liang2024alignment} which adapts Llama-3 checkpoints on Arabic, and the Arabic-English bilingual model \textbf{JAIS} \cite{sengupta2023jais}. We also experiment with encoder LMs such as \textbf{XLM-R} \cite{conneau-etal-2020-unsupervised}, as well as Arabic monolingual encoders, including \textbf{ARBERT} and \textbf{MARBERT} \cite{abdul2020arbert}, \textbf{CAMeLBERT} \cite{inoue2021interplay}, and \textbf{AraBERT} \cite{arabert}.

\begin{figure*}[t]
    \centering
    \includegraphics[width=\linewidth]{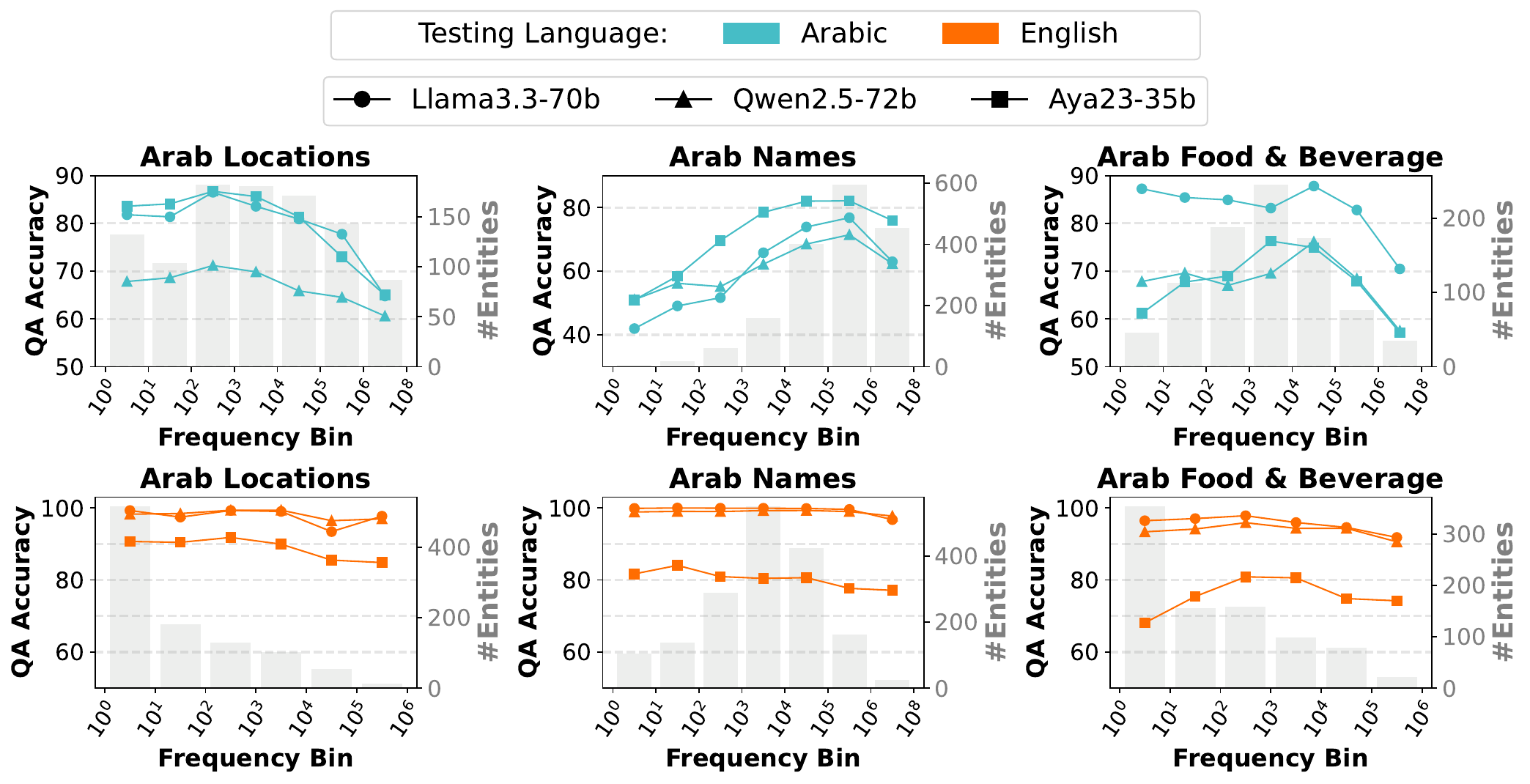}
    \caption{Average QA Accuracy ($\uparrow$) of LLMs when tested in Arabic and English on location, name, food, and beverage associated with Arab culture, stratified by their occurrence counts in the mC4 corpus (\S \ref{subsec:corpus-frequencies}; grouped into log10-spaced bins). Gray bars in background represent number of entities tested in each bin. Interestingly, LMs struggle with very high-frequency entities in Arabic.}
    \label{fig:frequency-results}
\end{figure*}

\subsection{Cultural Adaptation: Text Infilling}
\label{subsec:text-infilling-task}

We compare the ability of LMs at adapting to cultural contexts in Arabic and English by analyzing their preference for Arab vs. Western entities as \texttt{[MASK]} token fillings of CAMeL-2 contexts.

\paragraph{Text Infilling Setup.} We use the \textbf{C}ultural \textbf{B}ias \textbf{S}core (CBS) designed by \citet{naous-etal-2024-beer} as a likelihood-based measure of a LM's ability at adaptating to cultural contexts. Consider the sets of Arab entities $ A = \{a_i\}_{i=1}^{N}$ and Western entities $B = \{b_j\}_{j=1}^{M}$. The CBS for an Arab entity $a_i$ is the percentage of a LM's preference for Western entities when placed within the same context. For a set of culturally-grounded masked contexts $C = \{c_k\}_{k=1}^{K}$, we compute the $\mathrm{CBS}(a_i)$ as:

\begin{equation*}
      \frac{1}{K \times M}  \sum_{k=1}^K  \sum_{j=1}^{M} \mathbbm{1} [P_{\mathtt{[MASK]} }(b_j|t_k)  > P_{\mathtt{[MASK]} }(a_i|t_k) ]
\end{equation*}

where $P_{\texttt{[MASK]}}$ is the LM's probability of an entity filling the masked token. As the contexts are grounded in Arab culture (i.e., only Arab entities are appropriate), LMs are expected to score a CBS closer to 0\%. A higher CBS score indicates a stronger preference by LMs for Western entities in place of Arab entities, given the same context. For all entity types, we compute the CBS for a random sample of 50 Arab entities, where we test each against 50 randomly sampled Western entities.

\paragraph{Results.} Figure~\ref{fig:cbs-comparison-arabic-english} shows the average CBS per entity type for several LMs.  Interestingly, \textbf{LMs are better at adapting to Arab cultural contexts in English than in Arabic} with reduced CBS levels in nearly all cases, reaching the 15-30\% range for names and locations. One exception is food and beverage categories where a struggle is visible in both languages. We analyze the potential reasons behind these results in \S \ref{sec:origin-of-biases}.

\subsection{Cross-Cultural Performance}
\label{subsec:fairness-task}

Utilizing CAMeL-2, we compare the performance of LMs when tested in Arabic vs. English parallel contexts for extractive QA and NER tasks.

\paragraph{Prompting Setup for Extractive QA.} We evaluate GPT-type models in an extractive QA setup by prompting the LMs to extract the cultural entity from a given context in CAMeL-2 (see Appendix~\ref{appendix:extractive-qa-prompts} for prompts). We create an Arabic test set for each entity by replacing the \texttt{[MASK]} token of each context with the entity (i.e., $\sim$15 test contexts per entity), as well as a corresponding English test set with the same entities and contexts that are translations (\S \ref{subsec:enhancing-coverage}). We evaluate random samples of 1k Arab and 1k Western entities for each type and use all entities for types with less than 1k entities.

\paragraph{Fine-tuning Setup for NER.} We similarly evaluate BERT-type models on the NER task using the culturally-grounded contexts from CAMeL-2. We fine-tune models capable of recognizing \textit{names}, \textit{authors}, and \textit{locations} using the ANERCorp dataset for Arabic \cite{anercorp} and CoNLL-2003 for English  \cite{sang2003introduction}. For the remaining entity types (i.e., \textit{food}, \textit{beverage}, \textit{sports clubs}, \textit{religious places}) that are not covered by existing manually annotated NER corpora, we fine-tune LMs via distant supervision from Wikipedia articles \cite{liang2020bond} that we collected in \S \ref{subsec:enhancing-coverage} (see  Appendix~\ref{appendix:distant-supervision} for details). We exclude entities that appear in fine-tuning from our evaluations.

\paragraph{Results.} Table~\ref{tab:fairness-comparison-arabic-english} shows the average performance of Llama3.3-70b and XLMR$_{\mathrm{large}}$ on Arab and Western entities. We also show the performance difference ($\Delta$) between Western and Arab entities, where a positive $\Delta$ indicates better performance on Western entities. We find that \textbf{LMs have small performance gaps between the two cultures in English, but are consistently better at recognizing Western entities in Arabic than in English}, where differences reach up to 27 F1 points in NER and 15\% accuracy in extractive QA.

\section{On The Origin of Cultural Biases}
\label{sec:origin-of-biases}

Motivated by our observations in \S\ref{sec:bias-comparison-arabic-english}, we analyze various factors that may cause LMs to exhibit more severe Western bias in Arabic than in English. We first study the relationship between the frequency of entities in pre-training data and LM performance (\S\ref{subsec:corpus-frequencies}). Our findings lead to analysis of the impact of Arabic word polysemy on LM biases (\S\ref{subsec:entity-polysemy-analysis}). We also look at scenarios where entities exhibit lexical overlap with other languages that use the Arabic script (\S\ref{subsec:lexical-overlap}).  Finally, we examine the role that tokenization plays in the observed issues (\S\ref{subsec:tokenization-analysis}).

\subsection{Entity Frequency in Pre-training Data}
\label{subsec:corpus-frequencies}

We examine how LM performance on entities in extractive QA (\S\ref{subsec:fairness-task}) changes with respect to how often entities occur in pre-training. Since the pre-training corpora of state-of-the-art multilingual LMs are not public, we approximate the occurrence of entities in pre-training using the Arabic portion (0.96B lines) and English portion (3.1B lines) of the mC4 corpus\footnote{mC4 is a multilingual partition of \href{https://commoncrawl.org/}{CommonCrawl} web scrapes which are an essential part of LM pre-training data. Partitioning was done using the \href{https://github.com/google/cld3}{CLD3} language detector.} \cite{xue2021mt5}. Figure~\ref{fig:frequency-results} shows the average QA accuracy achieved by several LMs on Arab location, name, food, and beverage entities versus their occurrence count, which we group into log10-spaced bins (i.e., entities that appear 1 to 10 times, 10 to 100 times, etc.). We observe the following key findings:

\paragraph{LMs struggle on high-frequency entities.} There is a noticeable drop in LM performance on entities that appear at very high frequencies ($>$1M times in bin `$10^6$--$10^8$'). This drop is much steeper when LMs are tested in Arabic than English, and more prevalent since more Arab culture-associated entities appear at extremely high frequencies in Arabic. Similar trends are observed for other entity types and on the text-infilling task (see Appendix~\ref{appendix:entity-frequency-results}). Upon inspection of those entities, we find that many are Arabic words that can hold multiple senses in different contexts, besides being used to represent entities. We explore this impact of word polysemy more closely in \S\ref{subsec:entity-polysemy-analysis}. 

\paragraph{LMs also struggle with long-tail entities.}  We find that LMs can also perform poorly on long-tail entities which appear at low frequencies ($<$10 times in bin `$10^0$--$10^1$'). This is especially noticeable on names and food entities in Arabic, where performance improves gradually as entities appear more frequently. In general, we find LMs to perform the best in both languages on entities that appear at a medium frequency (e.g., 1k-100k range), but face difficulty with the edge cases (low and high-frequency).

\begin{figure}[t!]
    \centering  
    \includegraphics[width=0.9\linewidth]{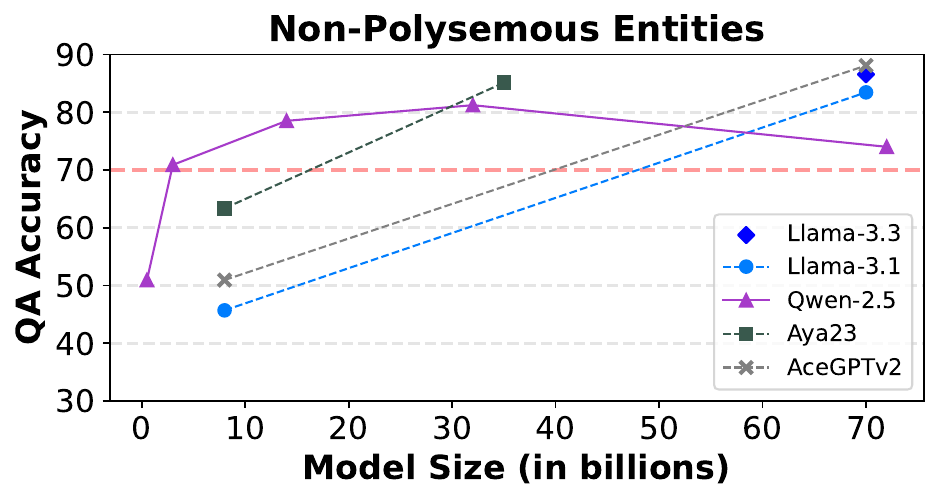}
    \includegraphics[width=0.9\linewidth]{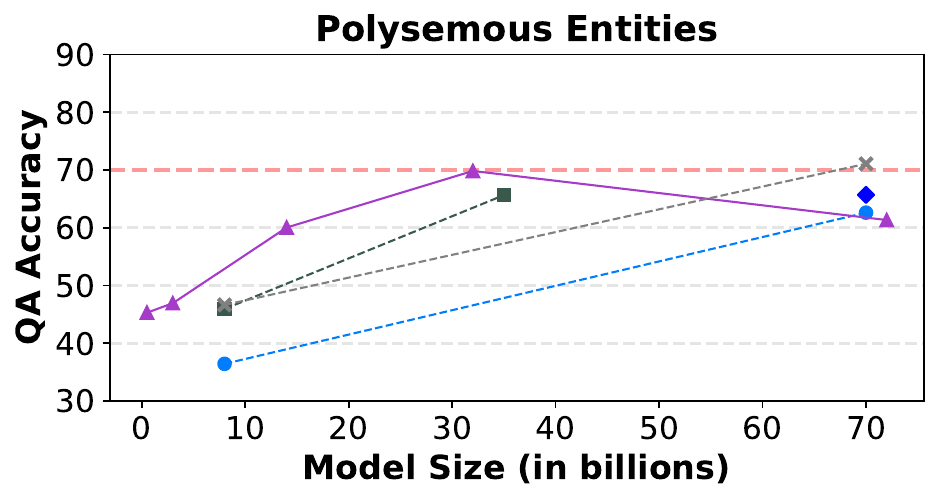}
    \caption{QA accuracy ($\uparrow$) for different sizes of LMs on high-frequency polysemous and non-polysemous Arab location entities. We find a positive scaling trend for all models, with lower performance on polysemous entities.}
    \label{fig:scaling-trend}
\end{figure}

\begin{figure*}[t]
    \centering
     \includegraphics[width=0.95\linewidth]{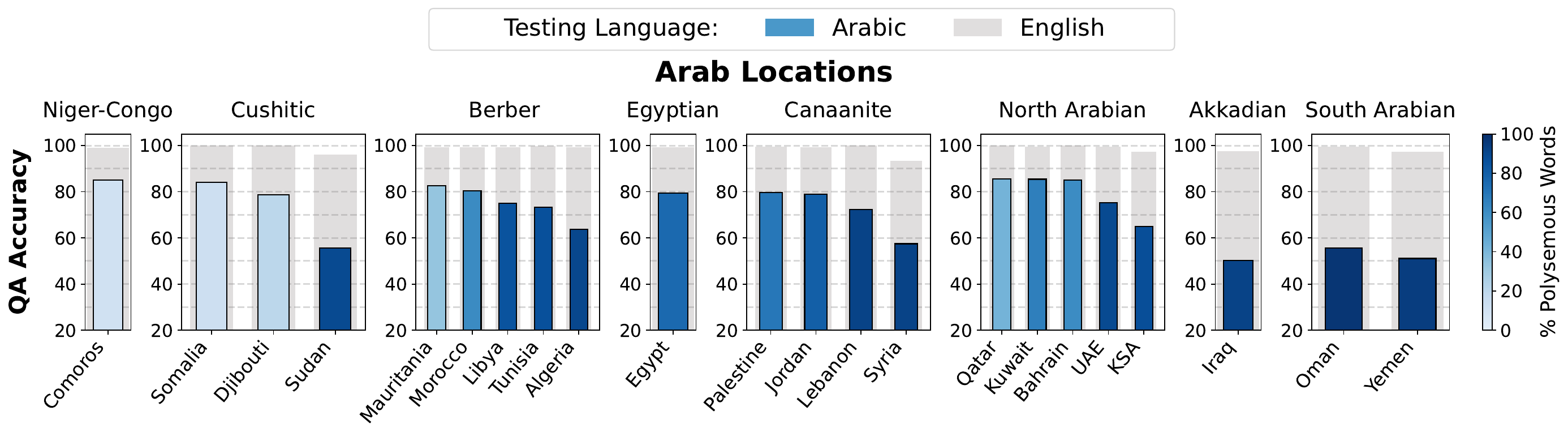}
    \includegraphics[width=0.95\linewidth]{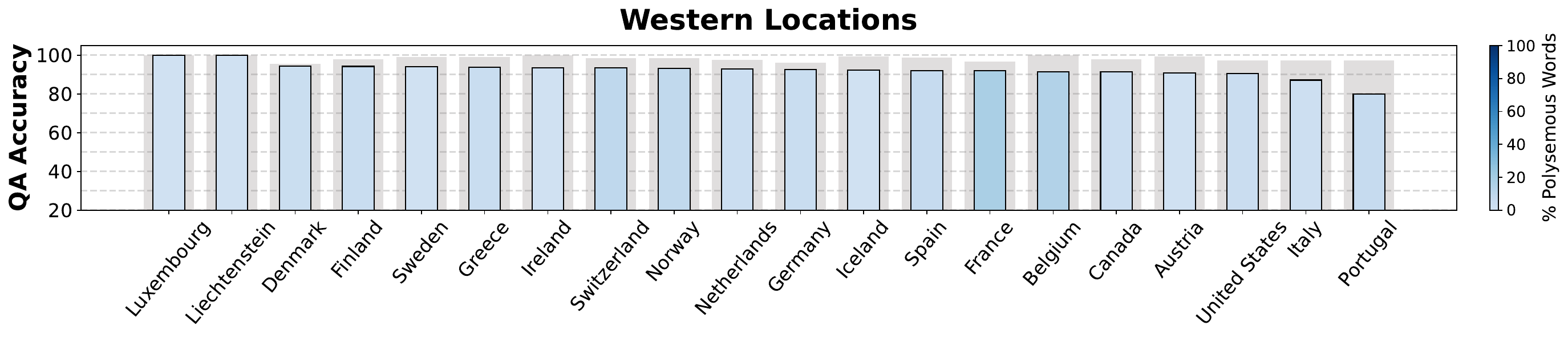}
    \caption{Average QA Accuracy ($\uparrow$) of Llama3.3-70b on the top-100 most frequent location entities in mC4 for each Arab and Western country in CAMeL-2 (\S \ref{subsec:entity-polysemy-analysis}). Arab countries are grouped by the language family that influences location naming in their region. Performance on Arab locations decreases as the percentage of entities that are Arabic polysemous words increases. Performance in English on the same entities is shown as a gray background.}
    \label{fig:performance-polysemy}
\end{figure*}

\subsection{Entities as Polysemous Words}
\label{subsec:entity-polysemy-analysis}

We further analyze how Arabic word polysemy impacts LM performance on entities that appear at high frequencies in pre-training corpora.

\paragraph{Background.} Consider the word \setcode{utf-8}``\<مطروحة>'' (pronounced: /ma-troo-ha/) as it appears in two Arabic sentences and their English translations:

\begin{enumerate}[(1)]
    \item  \<للنقاش> \underline{\<مطروحة>} \<القضية>  \\
    \textit{(The issue is \underline{proposed} for discussion)}  
    \item  \underline{\<مطروحة>} \<جدتي تسكن في> \\
    \textit{(My grandma lives in \underline{Matrooha})}
\end{enumerate}
 
In (1), the word appears in its literal sense ``\textit{proposed}". However, in (2) the same word is used as a noun to denote the name of a location. This dual use is common in Arabic, where the functionality of words used for entities changes depending on context. This is less common for entities in English, where there is generally a clear distinction between nouns and adjectives \cite{van2007theory}. We show this quantitatively in Appendix~\ref{appendix:comparing-polysemy}.

\paragraph{Setup.} We focus our analysis on location entities in Arab countries as they often exhibit regional linguistic influences. Thus, those entities can either be Arabic polysemous words or Arabic transliterations from languages that were historically spoken in those areas. For example, the current names of cities and villages in the Levant region could be Arabic transliterations from Canaanite languages that do not have other lexical uses in the Arabic language. This mixture of terms presents an ideal testing ground for LMs.\footnote{We refer the reader to Appendix~\ref{appendix:regional-influences} for more background on the regional linguistic influences in Arab countries.} We analyze QA performance on the 100 most frequent locations for each Arab country in the mC4 corpus. To determine if an entity matches an Arabic word that holds multiple meanings, we use the Almaany dictionary.\footnote{Almaany is a comprehensive resource that provides multiple meanings for Arabic words, highlighting their different uses across contexts: \url{https://www.almaany.com/}} We compare the performance with that of the 100 most frequent Western locations for each Western country in CAMeL-2. We use all locations available for Western countries with less than 100 locations.

\paragraph{Results.} Figure~\ref{fig:scaling-trend} shows the average QA accuracy for LMs of different sizes on the highly-frequent Arab locations, which we separate as polysemous and non-polysemous entities. In general, there is a positive scaling trend in both cases by all models. Yet, performance on polysemous entities is lower compared with non-polysemous ones, with 70B-sized models barely reaching the 70\% margin, as opposed to non-polysemous entities where most models reach near 90\%. 

The average results per country achieved by Llama3.3-70b are shown in Figure~\ref{fig:performance-polysemy}, where Arab countries are grouped by the influencing language on location naming in their region. We observe a trend where \textbf{QA performance in Arabic drops with the increase in the percentage of entities which are Arabic polysemous words}, with accuracy reducing drastically to the 40-60\% range. Performance is the best for countries where the percentage of polysemous words is low (e.g., Comoros, where entities are transliterations from the Comorian language). 

In contrast, we see that this issue is non-existent for Western entities in Arabic, where accuracy is near 90\% for all countries. This is because Western entities, being transliterations in Arabic, do not possess any other meaning. When the model is tested on the same entities in English, this problem also fades away, as Arab entities in English do not exhibit word polysemy either. This highlights that \textbf{the struggle of LMs with entities that are polysemous words in Arabic leads to a \textit{perceived} bias towards Western entities as they do not exhibit this phenomenon}. We also obtain similar results on NER with BERT-type LMs (see Appendix~\ref{appendix:polysemy-results}).

\begin{figure*}[t]
    \centering
    \includegraphics[width=\linewidth]{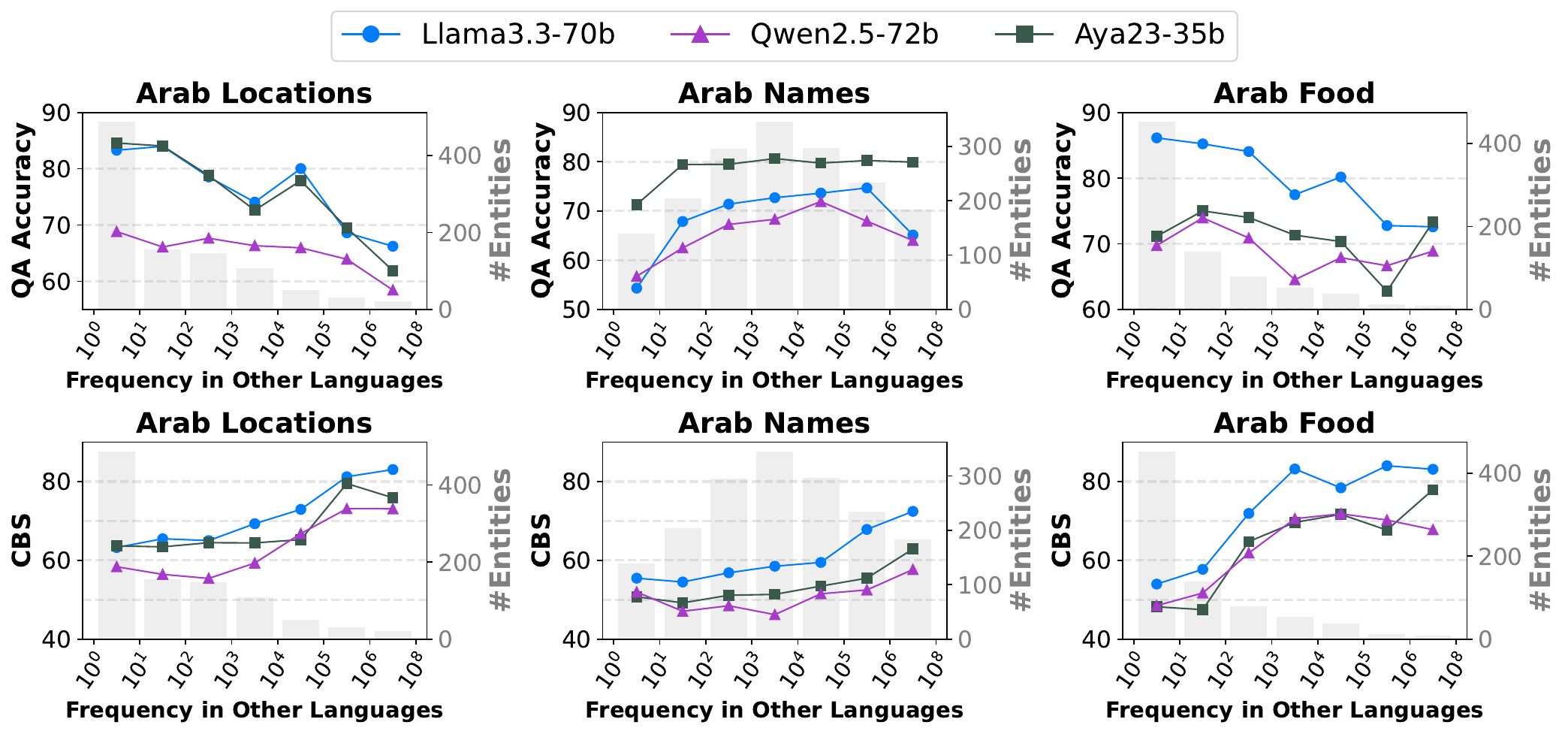}
    \caption{QA Accuracy ($\uparrow$)  and CBS ($\downarrow$) of LMs on Arab location, name, and food entities vs. their total count in other languages that use Arabic script (\textit{Farsi, Urdu, Tajik, Pashto, Kurdish}) in mC4 corpus. Performance decreases for all LMs as entities appear more frequently in other languages, especially for location and food entities.}
    \label{fig:frequency-results-other-langs}
\end{figure*}

\subsection{Other Languages Using Arabic Script} 
\label{subsec:lexical-overlap}

While Arabic script is primarily associated with the Arabic language, it is also used in several other languages, such as Farsi, Urdu, Kurdish, Tajik, and Pashto, due to historical and cultural connections between regions where they are spoken and the Arab world. There is thus a natural overlap of words between Arabic and those languages. We study how LMs behave on Arab entities as a function of their frequency in the pre-training data of other languages. We use the mC4 portions of those languages to obtain a total occurrence count for each Arab entity. 

\paragraph{Results.} Figure~\ref{fig:frequency-results-other-langs} shows the average QA accuracy and CBS at text-infilling achieved by LMs on Arab location, name, and food entities versus their total count in other languages. We observe a general trend where \textbf{LMs struggle on entities as they occur more frequently in other languages that share the script with Arabic}. Such entities occurring at very high frequencies can be common words in those languages that hold their own different meanings. For example, the word  \setcode{utf-8}``\<وزان>'' used to denote the Moroccan town ``\textit{Ouzanne}'' is also used in Farsi as the word for ``\textit{weight}''. 

As multilingual models are trained on those languages together, such lexical overlaps between entities in one language and highly frequent non-entity words in other languages with shared script can cause LMs to struggle at recognizing those entities. This issue is most noticeable for location and food entities but less so for name entities where performance is more stable. This could be due to the fact that Arab name entities are also used for first names in those languages, rather than having their own separate meanings.

\subsection{Subword Tokenization Matters}
\label{subsec:tokenization-analysis}

We analyze how tokenization \cite{kudo-2018-subword,song-etal-2021-fast,bostrom-durrett-2020-byte} impacts the behavior of LMs on entities. We compare the performance of LMs on NER and extractive QA with respect to how many tokens they get fragmented into. We also separate entities based on whether they exhibit polysemy at the token level, where we check if a tokens matches an Arabic word that can be used for different functionalities. 

\begin{figure*}[t]
    \centering
    \includegraphics[width=\linewidth]{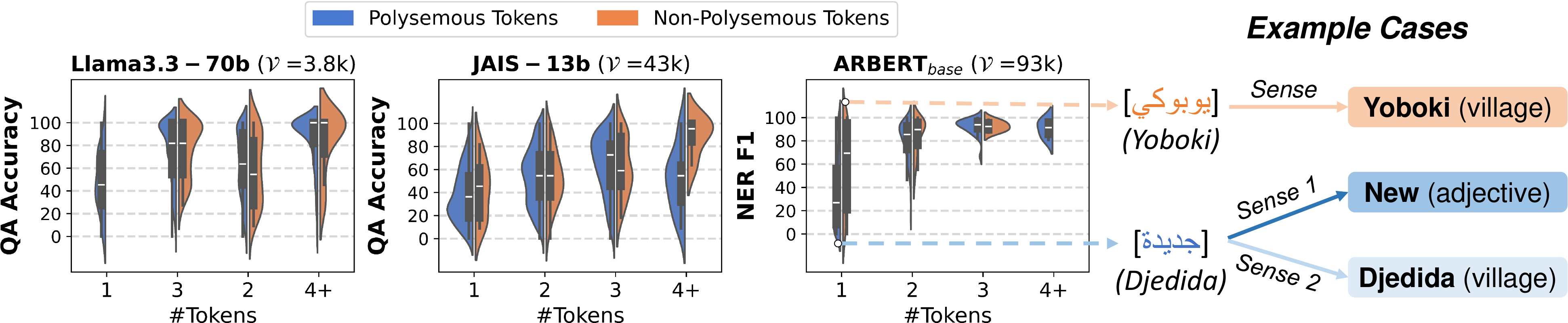}
    \caption{Performance distribution of Llama3.3-70b, JAIS-13b, and ARBERT on Arab location entities, in relation to how many tokens they get tokenized into. Entities are separated based on whether tokens correspond to Arabic polysemous words. $\mathcal{V}$ represents the number of Arabic tokens in each LM's vocabulary. Performance is the poorest on one-token entities that exhibit word polysemy, and improves on entities represented by multiple tokens.}
    \label{fig:tokenization-results}
\end{figure*}

\paragraph{Results.} Figure~\ref{fig:tokenization-results} shows the performance distribution on Arab location entities by Llama3.3-70b, JAIS-13b, and ARBERT. We find that \textbf{LMs perform the worst on entities that are tokenized into only one token, especially when it corresponds to a polysemous word}. Performance improves when entities are tokenized into multiple tokens, with gaps between entities containing polysemous and non-polysemous tokens gradually reducing. Interestingly, this issue is the most apparent for the ARBERT model, which is trained only on Arabic data and has a very large vocabulary of 93k tokens. 

\begin{figure}[t]
    \centering
    \includegraphics[width=\linewidth]{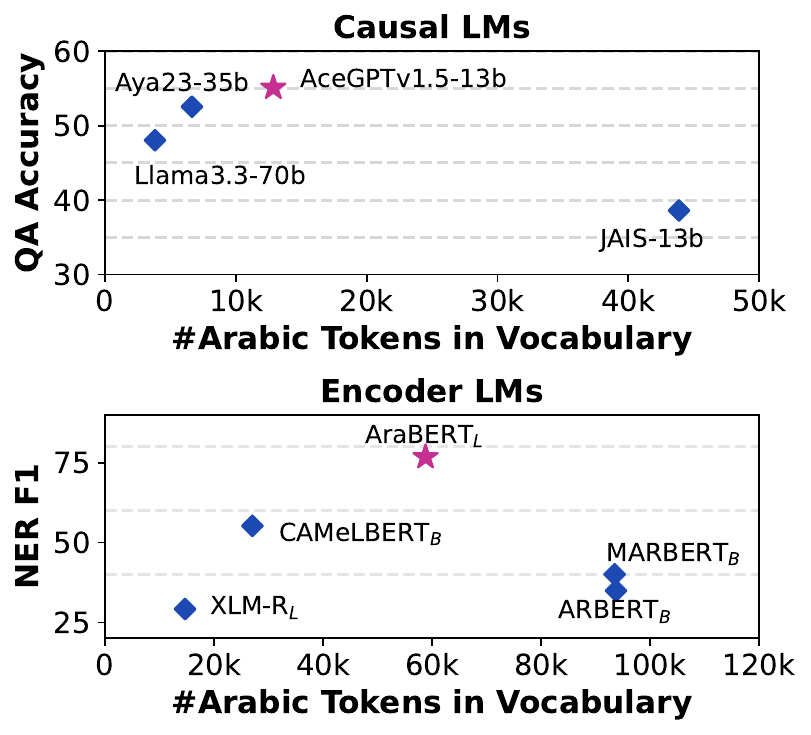} 
    \caption{Average QA Accuracy and F1 of LMs on one-token entities vs. the number of Arabic tokens in their vocabularies. Performance drops greatly as vocabularies get too large, with the best performing LMs (\includegraphics[height=0.7em]{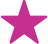}) having medium Arabic vocabulary sizes.}
    \label{fig:vocab-analysis}
\end{figure}

Figure~\ref{fig:vocab-analysis} shows the performance by LMs on one-token entities, in relation to the size of their Arabic vocabularies. We find that \textbf{LMs with medium Arabic vocabulary sizes perform the best, while performance drops for ones with very large vocabularies}. As the vocabulary size increases, frequency-based tokenization schemes will merge frequently used words into single tokens. This likely makes it more challenging for LMs to recognize entities in Arabic that exhibit word polysemy, as they get tokenized and encoded in the same way as when those words appear in text with non-entity senses within the same language or other languages sharing the same script. Such observations motivate the need for better tokenization approaches that can handle these cases for better cross-cultural performance.

\section{Related Work}

There has been growing interest in studying the cultural considerations surrounding LMs \cite{liu2024culturally, alkhamissi2024investigating}. This encompasses several aspects that LMs have to navigate such as \textit{cultural commonsense}, where LMs must differentiate between the societal norms (e.g., bringing a gift when visiting someone) of different cultures  \cite{shi2024culturebank, chiu2024culturalteaming, bhatt2024extrinsic, palta2023fork, fung2023normsage, huang2023culturally}. Other works explored \textit{culture-specific knowledge} in LMs (e.g., the color of the bridal dress)  and how their performance varies for different countries  \cite{keleg2023dlama, yin2022geomlama}. It has been shown that LMs are mostly familiar with a few cultures dominated in pre-training data, but struggle with less represented cultures \cite{shen2024understanding, rao2024normad, seth2024dosa}, and in non-English languages \cite{arora2024calmqa, masoud2023cultural}.

Another line of work explores the cultural appropriateness of LMs by analyzing their behavior when handling \textit{entities that exhibit cultural variation} such as first names \cite{an2024large, nghiem2024you, jeoung2023examining, an2023nichelle, gautam2024stop} and food dishes \cite{zhou2024does,li2024foodieqa}. The study of \citet{naous-etal-2024-beer} revealed performance gaps in LMs when handling entities associated with Arab vs. Western culture in Arabic language, where models were performed consistently better on Western entities. Our work builds on this study, with the aim of pinpointing the origins of such gaps. 

Past research on analyzing the cross-cultural performance of LMs have focused primarily on the representation of entities in pre-training data \cite{wolfe2021low, mukherjee2024global}, demonstrating a struggle of LMs to learn knowledge about entities that rarely appear in corpora \cite{li2024attributing, kandpal2023large}. However, those analyses have been restricted to the English language only. Different from prior work, we analyze multiple facets that can contribute to Western biases in LMs beyond pre-training data, focusing on cross-linguistic differences in phenomena exhibited by entities and the impact of subword tokenization.

\section{Conclusion}

We analyzed a variety of factors that can contribute to entity-related cultural biases in LMs. We showed how non-English linguistic phenomena such as word polysemy in Arabic, lexical overlaps with other languages, and frequency-based tokenization can cause performance degradation on entities associated with Arab culture, leading to perceived Western biases in LMs. We hope our study lays a foundation of important aspects to consider in building culturally fair multilingual LMs.

\section*{Limitations}

In this work, we investigated the origins of entity-related cultural biases in LMs by probing their ability at culturally-appropriate text-infilling and analyzing their cross-cultural performance on the extractive QA and NER tasks. There are other entity-related cultural biases that can also manifest in model behavior such as sentiment and stereotype associations in generated text \cite{naous-etal-2024-beer}. We leave the exploration of the reasons behind such learned associations for future work.

Our analyses showed that LMs struggle on entities in Arabic that are associated with Arab culture when they exhibit word polysemy. While Western entities transliterated into Arabic mostly do not exhibit this phenomenon, there are cases where a transliteration could match a random word used in Arabic language based on the closeness of their phonetic pronunciation. For example, the name ``\textit{Ben}'' could be transliterated as \setcode{utf-8}``\<بن>'' which is used in Arabic for ``\textit{son of}'' and ``\textit{powdered coffee}''. There are other cases where transliterations of less common entities could match the transliteration of other famous entities. For example, the name ``\textit{Yvonne}'' could be transliterated as \setcode{utf-8}``\<ايفون>'' which is also the common Arabic transliteration for ``\textit{iPhone}''. Such cases, while being rare, could cause LMs to fail at recognizing certain Western entities in Arabic. Although we did not explore the sensitivity of LMs to this phenomenon, the parallel entities in CAMeL-2 can offer a valuable resource for future studies to better analyze such cases.

Our work focuses on Arab culture and analyzes entity-related biases in Arabic language. Such biases may also be manifested by LMs in many other non-Western languages. Future studies can follow the process described in this work to extend CAMeL-2 to such languages.

\section*{Acknowledgements}
The authors would like to thank Tanmay Parekh and Kai-Wei Chang for valuable discussions; Nour Allah El Senary and Jad Matthew Bardawil for their help in data annotation; Jonathan Zheng, Anton Lavrouk, and Yao Dou for feedback on the draft.

\bibliography{references}

\clearpage
\newpage

\appendix

\section{CAMeL-2: Additional Details}
\label{appendix:camel2}

\begin{figure}
    \centering
    \includegraphics[width=\linewidth]{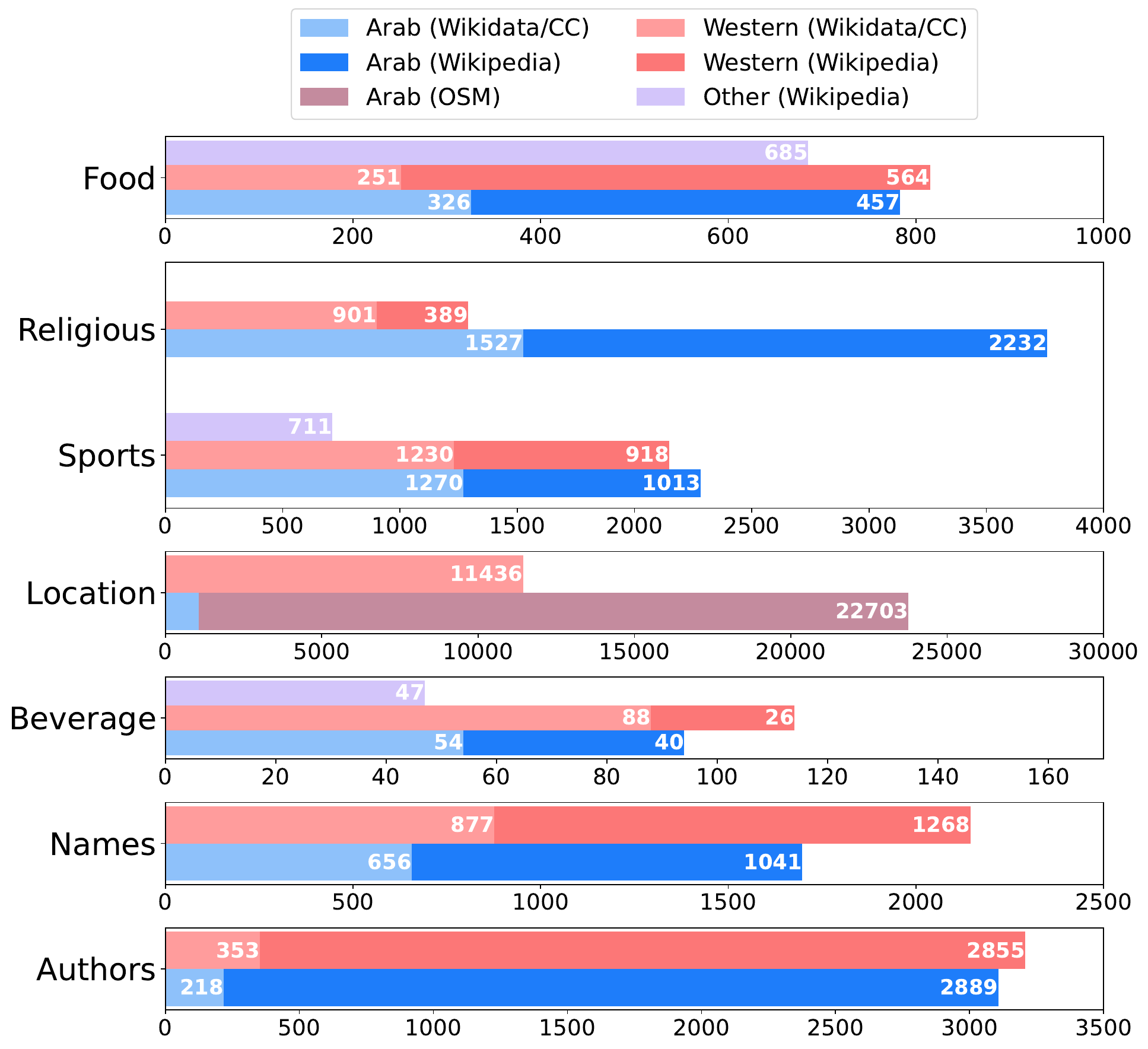}
    \caption{Distribution of entities in CAMeL-2 for each entity type stratified by association with \textit{Arab culture}, \textit{Western culture}, or \textit{Other Foreign Culture}, as well as their data collection source: Wikidata, CommonCrawl (CC), Wikipedia, OpenStreetMap (OSM).}
    \label{fig:entity-distribution-cultures}
\end{figure}

Figure~\ref{fig:entity-distribution-cultures} shows the distribution of entities in CAMeL-2 stratified by their association with \textit{Arab culture}, \textit{Western culture}, or \textit{Other Foreign Culture}, as well as their source of collection (Wikidata/CommonCrawl entities collected in CAMeL \cite{naous-etal-2024-beer}, and newly collected entities from Wikipedia and OpenStreetMap).

\begin{table}[t]
\centering
\begin{adjustbox}{width=\linewidth}
\begin{tabular}{@{}lll@{}}
\toprule
\multirow{2}{*}{\textbf{Entity Type}} & \multicolumn{2}{c}{\textbf{Wikipedia Categories}} \\
 & \multicolumn{1}{c}{\textit{Arabic}} & \multicolumn{1}{c}{\textit{English}} \\ \midrule
\multirow{2}{*}{Authors} & [adjective] \small{\<كتاب و كاتبات>} & [adjective] writers \\
 & [adjective] \small{\<روائيون>} & [adjective] authors \\ \midrule
Beverage & [adjective] \small{\<مطبخ>} & [adjective] cuisine \\ \midrule
Food & [adjective] \small{\<مطبخ>} & [adjective] cuisine \\ \midrule
\multirow{5}{*}{Names} & [adjective] \small{\<سياسيون>} & [adjective] politicians \\
 & [adjective] \small{\<رياضيون>} & [adjective] athletes \\
 & [adjective] \small{\<ممثلون>} & [adjective] actors \\
 & [adjective] \small{\<كتاب و كاتبات>} & [adjective] writers \\
 & [adjective] \small{\<روائيون>} & [adjective] authors \\ \midrule
\multirow{2}{*}{Religious} & [country] \small{\<مساجد في>} & mosques in [country] \\
 & [country] \small{\<كنائس في>} & churches in [country] \\ \midrule
Sports Clubs & [country] \small{\<أندية كرة قدم في>} & football clubs in [country] \\ \bottomrule
\end{tabular}
\end{adjustbox}
\caption{List of Arabic Wikipedia categories used to perform country-wise extraction of cultural entities. [adjective] refers to the country-specific adjective (e.g., \textit{Palestinian}, \textit{Irish}, \textit{Thai}, etc.).}
\label{tab:wikipedia-categories}
\end{table}

 \begin{table*}[t!]
\centering
\begin{adjustbox}{width=0.8\linewidth}
\begin{tabular}{@{}lcccccccccccc@{}}
\toprule
 & \multicolumn{6}{c}{\textbf{Aya23-35b}} & \multicolumn{6}{c}{\textbf{AceGPTv2-70b}} \\ \cmidrule(l){2-13} 
 & \multicolumn{3}{c}{\textbf{Arabic}} & \multicolumn{3}{c}{\textbf{English}} & \multicolumn{3}{c}{\textbf{Arabic}} & \multicolumn{3}{c}{\textbf{English}} \\ \cmidrule(l){2-13} 
 & \textit{Arab} & \textit{Western} & $\Delta$Acc & \textit{Arab} & \textit{Western} & $\Delta$Acc & \textit{Arab} & \textit{Western} & $\Delta$Acc & \textit{Arab} & \textit{Western} & $\Delta$Acc \\ \midrule
Authors & 91.68 & 88.96 & -2.72 & 79.03 & 81.73 & \cellcolor[HTML]{FFF3EA}2.70 & 89.43 & 82.86 & -6.57 & 88.61 & 95.50 & \cellcolor[HTML]{FFDDC4}6.89 \\
Beverage & 62.81 & \textbf{75.61} & \cellcolor[HTML]{80D2D8}12.80 & 76.47 & 76.06 & -0.41 & 89.47 & 81.92 & -7.55 & 98.62 & 96.06 & -2.56 \\
Food & 70.92 & 67.97 & -2.95 & 72.93 & 81.05 & \cellcolor[HTML]{FFDDC4}8.12 & 88.57 & 79.47 & -9.10 & 95.39 & 97.48 & \cellcolor[HTML]{FFF3EA}2.09 \\
Location & 81.04 & \textbf{91.55} & \cellcolor[HTML]{46BDC6}10.51 & 90.33 & 92.89 & \cellcolor[HTML]{FFF3EA}2.56 & 78.60 & \textbf{89.97} & \cellcolor[HTML]{80D2D8}11.37 & 97.88 & 98.96 & \cellcolor[HTML]{FFF3EA}1.08 \\
Names (F) & 78.78 & \textbf{87.30} & \cellcolor[HTML]{80D2D8}8.52 & 89.85 & 85.91 & -3.94 & 77.73 & \textbf{88.78} & \cellcolor[HTML]{80D2D8}11.05 & 99.84 & 99.09 & -0.75 \\
Names (M) & 79.53 & \textbf{80.70} & \cellcolor[HTML]{DCF3F4}1.17 & 72.96 & 73.90 & \cellcolor[HTML]{FFF3EA}0.94 & 85.03 & \textbf{87.83} & \cellcolor[HTML]{DCF3F4}2.80 & 94.53 & 94.45 & -0.08 \\
Sports & 47.88 & \textbf{53.85} & \cellcolor[HTML]{80D2D8}5.97 & 81.78 & 77.76 & -4.02 & 72.20 & 66.34 & -5.86 & 80.24 & 83.37 & \cellcolor[HTML]{FFF3EA}3.13 \\
Religious & 82.49 & 79.35 & -3.14 & 90.11 & 93.47 & 3.36 & 80.07 & \textbf{81.65} & \cellcolor[HTML]{DCF3F4}1.58 & 79.29 & 84.67 & \cellcolor[HTML]{FFDDC4}5.38 \\ \bottomrule
\end{tabular}
\end{adjustbox}
\caption{Average QA Accuracy of Aya23-35b and AceGPTv2-70b on Arab and Western entities when tested in Arabic and English. $\Delta$Acc represents performance differences between Western and Arab entities.}
\label{tab:fairness-other-models}
\end{table*}

\paragraph{Wikipedia-based Extraction.} The Arabic Wikipedia categories used to perform country-wise extraction of articles for each entity type are listed in Table~\ref{tab:wikipedia-categories}. We first de-duplicate the extracted articles from Wikipedia, as many of them would be cross-listed under the category of multiple countries. Many of the extracted articles would be irrelevant to the entity type of interest and were thus manually filtered out. For example, in addition to articles about food dishes, the ``\texttt{[country adjective] cuisine}" category would also contain articles about particular chefs or restaurants in that country. We finally inspect the titles of the remaining articles and remove any additional text between parentheses that is not part of the entity (e.g., a title such as ``\textit{Mandi (food)}" where \textit{(food)} was manually removed).

For author names, the number of articles in the \texttt{authors} category was very large. We thus took a random sample of 3k articles from Arab countries and 3k articles from Western countries that were then manually filtered by the annotators.

To collect first names, we first extract all article titles and text from multiple Wikipedia categories, as shown in Table~\ref{tab:wikipedia-categories}, that relate to human entities (\textit{politicians}, \textit{actors}, \textit{athletes}, \textit{writers}, and \textit{authors}). We then extract the first uni-gram and bi-gram from each article title (which represents the full name of the human entity) and perform de-duplication. Our annotators then filter out extractions that are not person names and classify the extracted names for cultural association (\textit{Arab}, \textit{Westerm}, or \textit{Other Foreign Culture}). The extracted names were also classified as masculine or feminine by the annotators, a step that is necessary to match the gendered grammar of the Arabic language. The annotators were also given access to the corresponding Wikipedia article for each extraction to guide their decision in annotation. This process was done for all Wikipedia articles extracted from Arab countries, resulting in 1,268 new Arab names. The articles obtained from Western countries were too large in number (> 20k articles), we thus took a random sample of 1.5k articles that were filtered by the annotators.

\paragraph{Location Extraction from Georgraphic Data.} Open Street Maps (OSM) uses a ``\texttt{place}'' tag to represent various types of locations. For each Arab country, we extract all locations that have place tags of \textit{city}, \textit{town}, \textit{village}, \textit{neighborhood}, and \textit{suburb}. We discard locations that have other highly-specific place tags such as \textit{isolated dwelling}, \textit{hamlet}, \textit{farm}, etc. since these represent individual residential structures, often in remote areas, rather than organized settlements with social significance. We also discarded locations which are more than one word expressions as they mostly consisted of repeatedly used terms (e.g., hill, mountain, valley, spring, etc.).

\begin{figure}[t]
    \centering
    \includegraphics[width=\linewidth]{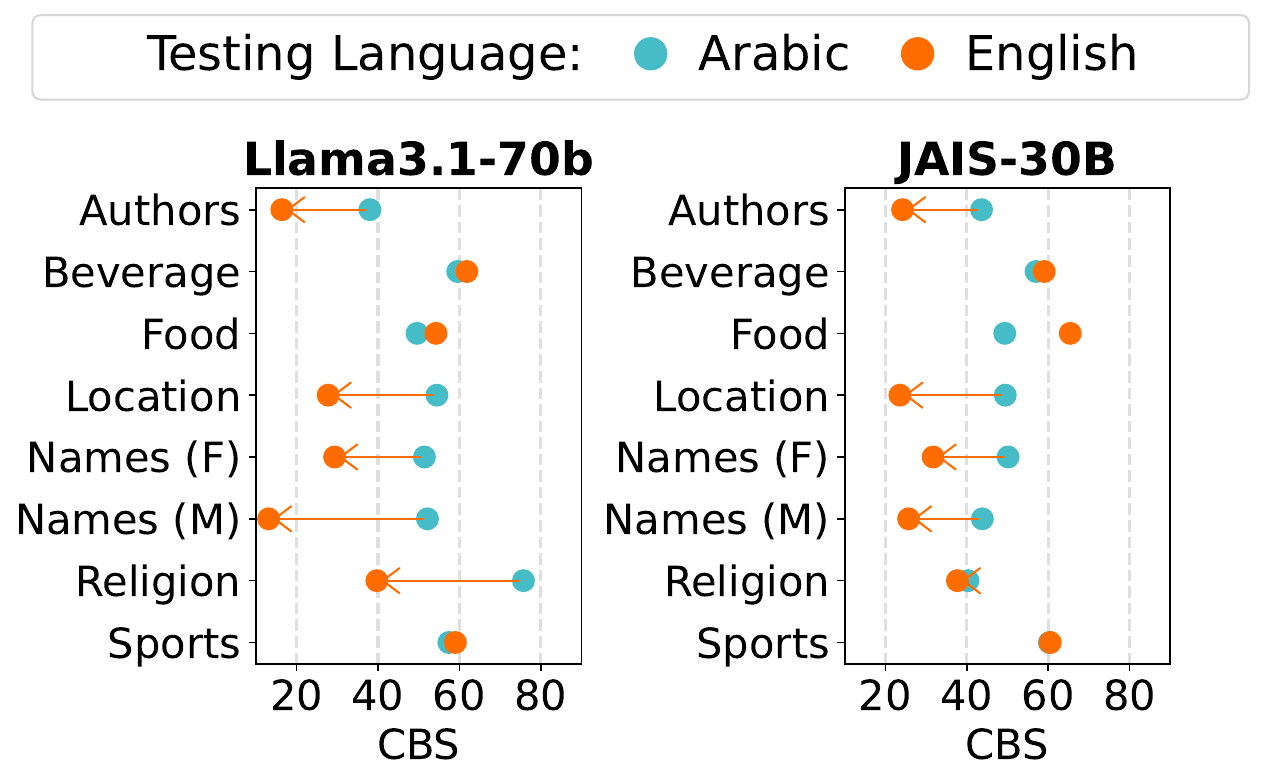}
    \caption{Average CBS per entity type achieved by Llama3.1-70b and JAIS-30b on culturally-grounded contexts from CAMeL-2.}
    \label{fig:cbs-ar-en-additional}
\end{figure}

\section{Arabic vs. English Comparisons}
\label{appendix:arabic-vs-english-comparison}

\subsection{Text-Infilling}
\label{appendix:cultural-bias-score}

\paragraph{Additional Results.} Figure~\ref{fig:cbs-ar-en-additional} reports CBS results on text-infilling by additional Llama3.1-70b and JAIS-30b when tested in both Arabic and English. Similar to our observations in \S\ref{subsec:text-infilling-task}, we see a consistent trend across all LMs where better adaptation to Arab cultural contexts is achieved in English compared to Arabic, where CBS values are greatly reduced.

\begin{figure}[t]
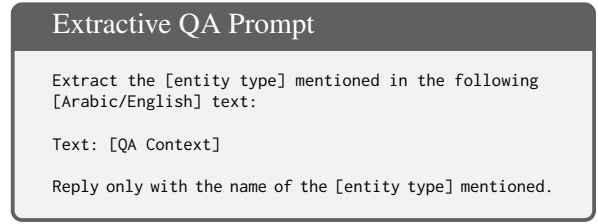

\centering
\begin{tcolorbox}
[colback=black!5!white,colframe=gray!75!black,title=Extractive QA Prompt]
\scriptsize
\begin{verbatim}
Extract the [entity type] mentioned in the following
[Arabic/English] text:

Text: [QA Context]

Reply only with the name of the [entity type] mentioned.
\end{verbatim}
\end{tcolorbox}
\caption{Prompt template used perform extractive QA with GPT-type LMs. }
\label{fig:qa-prompt}
\end{figure}

\subsection{Extractive QA}
\label{appendix:extractive-qa-prompts}

\paragraph{Prompt.} The prompt template used for performing extractive QA of entities with GPT-type LMs is shown in Figure~\ref{fig:qa-prompt}. The \texttt{[entity type]} in the template is replaced with the name of the entity type of interest (i.e., location, person's name, author name, food dish, drink,  mosque, church, football club). The \texttt{[QA Context]} is replaced by one of the QA contexts collected for the respective entity type (\S\ref{subsec:natural-contexts}) where the entity to be extracted is placed instead of the \texttt{[MASK]} token of the context. The instruction is given in English for all models tested, expect for JAIS where giving the model the instruction in Arabic lead to better performance.

\paragraph{Additional Results.} Table~\ref{tab:fairness-other-models} shows additional results achieved by Aya23-35b and AceGPTv2-70b on Arab and Western entities across different entity types. We observe similar results to those achieved by Llama3.3-70b (\S~\ref{subsec:fairness-task}) where we see large performance gaps between Western and Arab entities in Arabic for most entity types, but gaps between cultures are much smaller in English. Notably, we do find an improvements in those models on Arab entities in the religious and food entity type compared with Llama3.3-70b, which may be due to additional training on Arabic data.

\begin{table}[t]
\centering
\begin{adjustbox}{width=\linewidth}
\begin{tabular}{@{}lcc@{}}
\toprule
\multirow{2}{*}{\textbf{Entity Type}} & \multicolumn{2}{c}{\textbf{\# Fine-tuning Samples} (train/val/test)} \\ \cmidrule(l){2-3} 
 & \textit{Arabic} & \textit{English} \\ \midrule
Food \& Beverage & 6,908/768/853 & 27,635/3,071/1,617 \\ \midrule
Religious & 2,655/295/328 & 12,623/1,403/1,559 \\ \midrule
Sports Clubs & 2,484/277/307 & 18,585/2,066/2,295 \\ \bottomrule
\end{tabular}
\end{adjustbox}
\caption{Number of NER fine-tuning examples (as train/val/test splits) constructed automatically from Wikipedia articles for \textit{food}, \textit{beverage}, \textit{sports clubs}, and \textit{religious place} entities. }
\label{tab:finetuning-splits}
\end{table}

\begin{table}[t]
\centering
\begin{adjustbox}{width=\linewidth}
\begin{tabular}{@{}lcccccc@{}}
\toprule
\multirow{3}{*}{\textbf{Model}} & \multicolumn{6}{c}{\textbf{Test Set F1 Score}} \\ \cmidrule(l){2-7} 
 & \multicolumn{3}{c}{\textit{Arabic}} & \multicolumn{3}{c}{\textit{English}} \\ \cmidrule(l){2-7} 
 & Foo & Spo & Rel & Foo & Spo & Rel \\ \midrule
XLMR$_{\mathrm{large}}$ & 77.32 & 85.62 & 91.52 & 90.47 & 82.30 & 82.26 \\
XLMV$_{\mathrm{base}}$ & 76.39 & 87.14 & 90.73 & 91.75 & 83.17 & 80.63 \\
ARBERT & 79.44 & 86.48 & 92.43 & --- & --- & --- \\ \bottomrule
\end{tabular}
\end{adjustbox}
\caption{F1 score achieved by various BERT-type LMs on the test set of the fine-tuning samples created automatically from Wikipedia.}
\label{tab:test-set-perf}
\end{table}

\subsection{Cultural Fairness: NER}
\label{appendix:distant-supervision}

\paragraph{Distant Supervision.} To fine-tune LMs for NER of \textit{food}, \textit{beverage}, \textit{sports clubs}, and \textit{religious places}, we leverage text from Wikipedia articles to perform distantly supervised fine-tuning. Specifically, we use entities from CAMeL-2 that are linked to Wikipedia articles. For each of these entities, we automatically create fine-tuning samples by extracting sentences from their corresponding Wikipedia articles where the entity is mentioned. We exclude entities from fine-tuning that do not appear on Wikipedia (i.e., entities extracted from Wikidata or Commoncrawl in CAMeL \cite{naous-etal-2024-beer}). Table~\ref{tab:finetuning-splits} shows the number of fine-tuning examples created in both Arabic and English.

\paragraph{Additional Results.} Table~\ref{tab:fairness-xlmv} shows the average F1 achieved by XLMV$_{\mathrm{base}}$ on Arab and Western entities across different entity types when tested in Arabic and English. Table~\ref{tab:fairness-arbert} shows the results of the Arabic monolingual ARBERT model when tested in Arabic. We observe similar trends to our analysis in (\S~\ref{subsec:fairness-task}) where BERT-type LMs are consistently better at recognizing Western entities in the Arabic language, but performance gaps between cultures in much smaller when LMs are tested in English.

\begin{table}[t]
\centering
\begin{adjustbox}{width=0.95\linewidth}
\begin{tabular}{@{}lccc!{\vrule}ccc@{}}
\toprule
 & \multicolumn{6}{c}{\textbf{XLMV$_{\mathrm{base}}$}} \\ \cmidrule(l){2-7} 
 & \multicolumn{3}{c}{\textbf{Arabic}} & \multicolumn{3}{c}{\textbf{English}} \\ \cmidrule(l){2-7} 
 & \textit{Arab} & \textit{Western} & $\Delta$F1 & \textit{Arab} & \textit{Western} & $\Delta$F1 \\ \midrule
Authors & 93.14 & \textbf{94.99} & \cellcolor[HTML]{DCF3F4}1.85 & 96.61 & 95.13 & -1.48 \\
Beverage & 52.87 & \textbf{64.58} & \cellcolor[HTML]{46BDC6}11.71 & 95.06 & 88.48 & -6.58 \\
Food & 50.15 & \textbf{62.16} & \cellcolor[HTML]{46BDC6}12.01 & 92.51 & 90.55 & -1.96 \\
Location & 64.55 & \textbf{89.27} & \cellcolor[HTML]{46BDC6}24.72 & 92.42 & 97.65 & \cellcolor[HTML]{FFDDC4}5.23 \\
Names (F) & 48.48 & \textbf{75.68} & \cellcolor[HTML]{46BDC6}27.20 & 98.32 & 97.57 & -0.75 \\
Names (M) & 70.37 & \textbf{80.52} & \cellcolor[HTML]{46BDC6}10.15 & 93.84 & 91.66 & -2.18 \\
Sports & 75.37 & \textbf{83.68} & \cellcolor[HTML]{98DBE0}8.31 & 91.83 & 92.68 & \cellcolor[HTML]{FFF3EA}0.85 \\
Religious & 71.22 & \textbf{84.23} & \cellcolor[HTML]{46BDC6}13.01 & 90.61 & 86.06 & -4.55 \\ \bottomrule
\end{tabular}
\end{adjustbox}
\caption{Average F1 of XLMV$_{\mathrm{base}}$ \cite{liang2023xlm} on Arab and Western entities when tested in Arabic and English. $\Delta$F1 represents performance differences between Western and Arab entities.}
\label{tab:fairness-xlmv}
\end{table}

\begin{table}[t]
\centering
\begin{adjustbox}{width=0.6\linewidth}
\begin{tabular}{@{}lccc@{}}
\toprule
 & \multicolumn{3}{c}{\textbf{ARBERT}} \\ \cmidrule(l){2-4} 
 & \multicolumn{3}{c}{\textbf{Arabic}} \\ \cmidrule(l){2-4} 
 & \textit{Arab} & \textit{Western} & $\Delta$F1 \\ \midrule
Authors & 96.01 & 93.83 & -2.18 \\
Beverage & 53.35 & \textbf{65.05} & \cellcolor[HTML]{46BDC6}11.70 \\
Food & 52.01 & \textbf{64.24} & \cellcolor[HTML]{46BDC6}12.23 \\
Location & 60.47 & \textbf{89.32} & \cellcolor[HTML]{46BDC6}28.85 \\
Names (F) & 69.61 & \textbf{75.81} & \cellcolor[HTML]{80D2D8}6.21 \\
Names (M) & 83.82 & \textbf{84.86} & \cellcolor[HTML]{DCF3F4}{\color[HTML]{333333} 1.04} \\
Sports & \multicolumn{1}{l}{90.81} & \textbf{94.66} & \cellcolor[HTML]{DCF3F4}3.85 \\
Religious & 59.48 & \textbf{69.08} & \cellcolor[HTML]{46BDC6}9.60 \\ \bottomrule
\end{tabular}
\end{adjustbox}
\caption{Average F1 of the Arabic monolingual ARBERT model on Arab and Western entities when tested in Arabic and English. $\Delta$F1 represents performance differences between Western and Arab entities.}
\label{tab:fairness-arbert}
\end{table}

\subsection{Experimental Details}

\paragraph{Language Models.} Table~\ref{tab:lms-repos} lists all the LMs used in our experiments, including the varying sizes used in our scaling analysis (\S\ref{subsec:entity-polysemy-analysis}) and different versions of the models for our vocabulary size analysis (\S\ref{subsec:tokenization-analysis} and Appendix~\ref{appendix:tokenization-results}).

\begin{table}[t]
\begin{adjustbox}{width=\linewidth}
\begin{tabular}{@{}ll@{}}
\toprule
\textbf{Language Model} & \textbf{Hugging Face Repository} \\ \midrule
\multicolumn{2}{c}{\textit{Causal LMs}} \\
Llama3.3-70b & \href{https://huggingface.co/meta-llama/Llama-3.3-70B-Instruct}{meta-llama/Llama-3.3-70B-Instruct} \\
Llama3.1-8b & \href{https://huggingface.co/meta-llama/Meta-Llama-3.1-8B-Instruct}{meta-llama/Meta-Llama-3.1-8B-Instruct} \\
Llama3.1-70b & \href{https://huggingface.co/meta-llama/Meta-Llama-3.1-70B-Instruct}{meta-llama/Meta-Llama-3.1-70B-Instruct} \\
Aya23-8b & \href{https://huggingface.co/CohereForAI/aya-23-8B}{CohereForAI/aya-23-8B} \\
Aya23-35b & \href{https://huggingface.co/CohereForAI/aya-23-35B}{CohereForAI/aya-23-35B} \\
Qwen2.5-0.5b & \href{https://huggingface.co/Qwen/Qwen2.5-0.5B-Instruct}{Qwen/Qwen2.5-0.5B-Instruct} \\
Qwen2.5-3b & \href{https://huggingface.co/Qwen/Qwen2.5-3B-Instruct}{Qwen/Qwen2.5-3B-Instruct} \\
Qwen2.5-14b & \href{https://huggingface.co/Qwen/Qwen2.5-14B-Instruct}{Qwen/Qwen2.5-14B-Instruct} \\
Qwen2.5-32b & \href{https://huggingface.co/Qwen/Qwen2.5-32B-Instruct}{Qwen/Qwen2.5-32B-Instruct} \\
Qwen2.5-72b & \href{https://huggingface.co/Qwen/Qwen2.5-72B-Instruct}{Qwen/Qwen2.5-72B-Instruct} \\
AceGPTv1.5-13b & \href{https://huggingface.co/FreedomIntelligence/AceGPT-v1.5-13B-Chat}{FreedomIntelligence/AceGPT-v1.5-13B-Chat} \\
AceGPTv2-8b & \href{https://huggingface.co/FreedomIntelligence/AceGPT-v2-8B-Chat}{FreedomIntelligence/AceGPT-v2-8B-Chat} \\
AceGPTv2-70b & \href{https://huggingface.co/FreedomIntelligence/AceGPT-v2-70B-Chat}{FreedomIntelligence/AceGPT-v2-70B-Chat} \\
JAIS-13b & \href{https://huggingface.co/inceptionai/jais-13b-chat}{inceptionai/jais-13b-chat} \\
JAIS-30b & \href{https://huggingface.co/inceptionai/jais-13b-chat}{inceptionai/jais-30b-chat-v3} \\ \midrule
\multicolumn{2}{c}{\textit{Encoder LMs (multilingual)}} \\
XLMR$_{\mathrm{large}}$ & \href{https://huggingface.co/FacebookAI/xlm-roberta-large}{FacebookAI/xlm-roberta-large} \\
XLMV$_{\mathrm{base}}$ & \href{https://huggingface.co/facebook/xlm-v-base}{facebook/xlm-v-base} \\
mDeBERTa-v3$_{\mathrm{base}}$ & \href{https://huggingface.co/microsoft/mdeberta-v3-base}{microsoft/mdeberta-v3-base} \\
GigaBERT$_{\mathrm{base}}$ & \href{https://huggingface.co/lanwuwei/GigaBERT-v3-Arabic-and-English}{lanwuwei/GigaBERT-v3-Arabic-and-English} \\ \midrule
\multicolumn{2}{c}{\textit{Encoder LMs (monolingual - Arabic only)}} \\
AraBERT$_{\mathrm{large}}$ & \href{https://huggingface.co/aubmindlab/bert-large-arabertv02}{aubmindlab/bert-large-arabertv02} \\
ARBERT$_{\mathrm{base}}$ & \href{https://huggingface.co/UBC-NLP/ARBERT}{UBC-NLP/ARBERT} \\
MARBERT$_{\mathrm{base}}$ & \href{https://huggingface.co/UBC-NLP/MARBERT}{UBC-NLP/MARBERT} \\
CAMeLBERT$_{\mathrm{base}}$ & \href{https://huggingface.co/CAMeL-Lab/bert-base-arabic-camelbert-mix}{CAMeL-Lab/bert-base-arabic-camelbert-mix} \\ \bottomrule
\end{tabular}
\end{adjustbox}
\caption{List of LMs used in our experiments and their repository links on Hugging Face.}
\label{tab:lms-repos}
\end{table}

\paragraph{QA Inference.} We ran our experiments using 8 NVIDIA A40 GPUs. For inference on the extractive QA task with causal LMs, we used the vLLM library\footnote{\url{https://docs.vllm.ai}} \cite{kwon2023efficient} for fast inference. We performed greedy decoding by setting the following parameters \{\texttt{temperature=0}, \texttt{top\_p=1}, \texttt{top\_k=1}\}. We also limit the number of generated tokens by the models by setting \{\texttt{max\_tokens=30}\}. Futher, we limit the context length by setting \{\texttt{max\_model\_len=4096}\}.

\paragraph{Fine-tuning.} For fine-tuning BERT-type models on the NER task, we fine-tuned each model for 5 epochs using the cross-entropy loss and the Adam optimizer and tuned the learning rate in the set $\{1e^{-5}, 1e^{-6}, 1e^{-7}\}$. We selected checkpoints based on the best F1 on the validation set. Fine-tuning was done using one NVIDIA A100 GPU.

\paragraph{Entity Occurrence Counts.} To obtain counts of entities in the mC4 corpus, we use the Aho-Corasick string search algorithm\footnote{\url{https://pyahocorasick.readthedocs.io}} where we construct finite state machines using the entities, allowing for efficient transversal over the corpora.

 \begin{figure*}[t]
    \centering
    \includegraphics[width=0.32\linewidth]{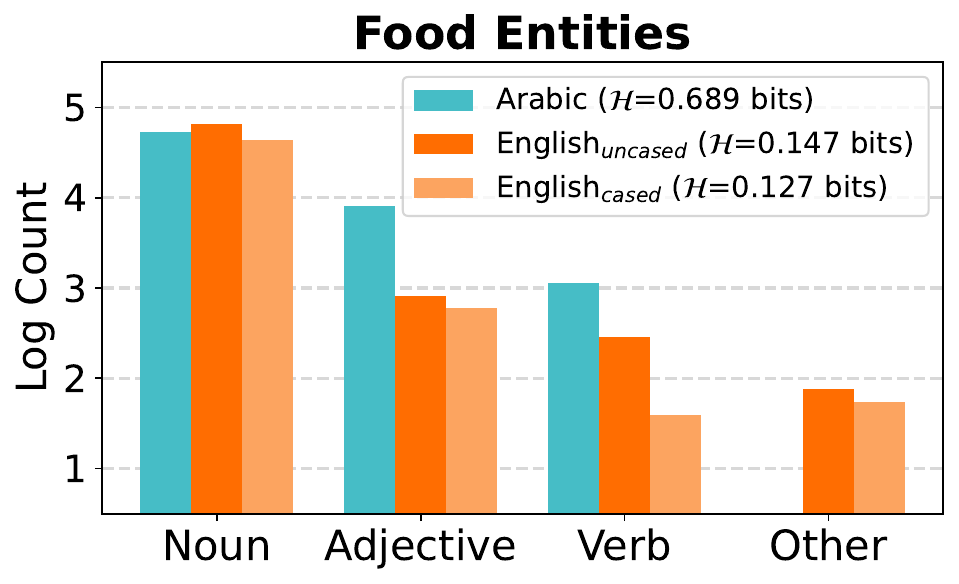}
    \includegraphics[width=0.32\linewidth]{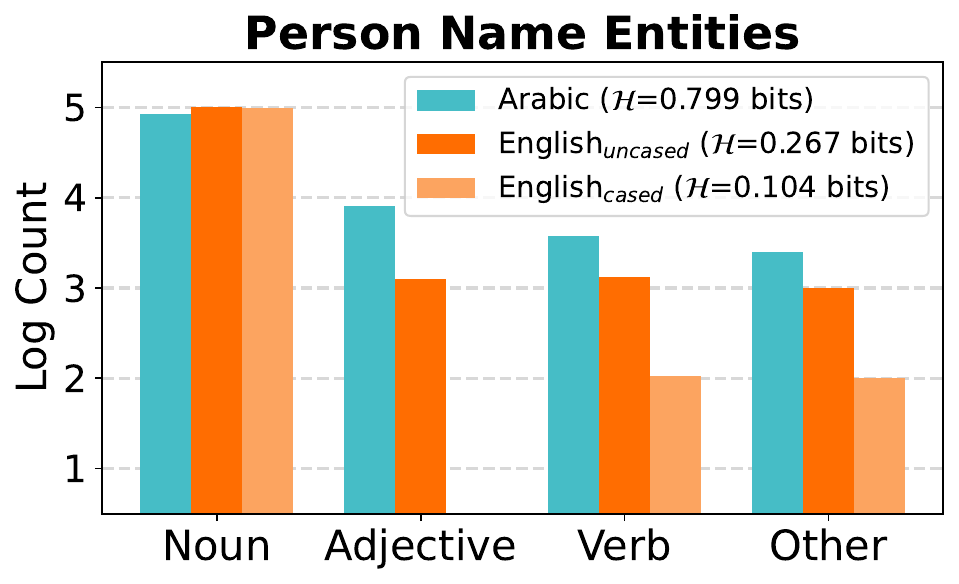}
    \includegraphics[width=0.32\linewidth]{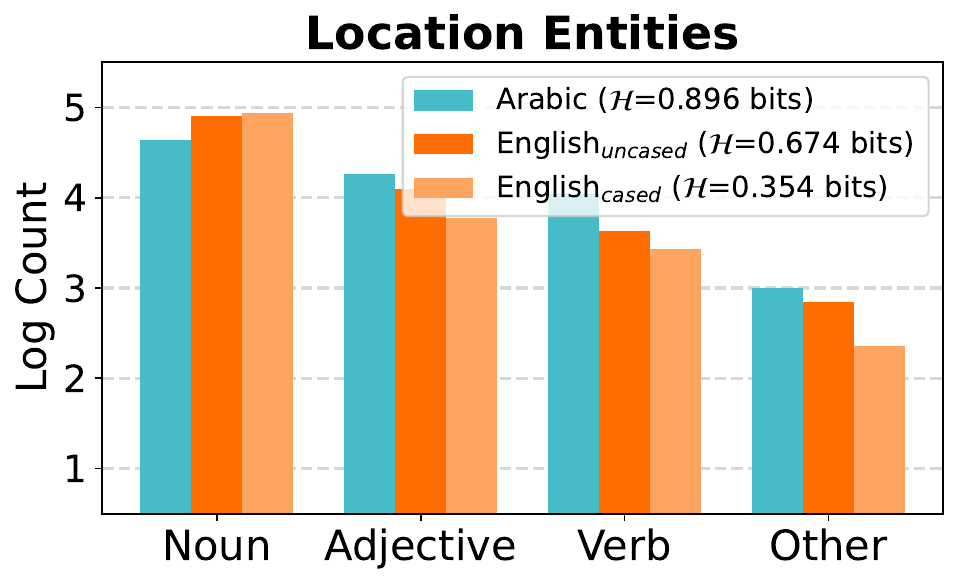}
    \caption{POS tag distribution of the 100 most frequent food, name, and location entities in Arabic and English mC4. Entities in Arabic encode higher information as they appear more often with different grammatical roles.}
    \label{fig:entity-information-pos}
\end{figure*}

\section{Arabic Entities as Polysemous Words}

\subsection{Comparing Polysemy Across Languages}
\label{appendix:comparing-polysemy}

To quantify and compare the prevalence of entity polysemy in Arabic and English, we analyze how often entities are used as different parts of speech (such as nouns, verbs, or adjectives) in texts written in both languages. We use the first 10M documents from the Arabic and English portions of the mC4 pre-training corpus \cite{raffel2020exploring}, which we then tokenize into sentences, yielding 239M Arabic sentences and 209M English sentences. Using the Arabic and English entities from CAMeL-2, we then identify the top 100 most frequent name, food, and location entities in the corpora. For each entity, we randomly sample 1000 sentences in which they appear, then determine their part-of-speech tag in each sentence using the Farasa POS tagger \cite{darwish2016farasa} for Arabic and the Stanford POS tagger \cite{qi2018universal} for English. We perform this analysis for English entities both with and without uppercasing of the first letter.

Figure~\ref{fig:entity-information-pos} shows the distribution of part-of-speech tags for the 100 most frequent name and food entities in Arabic and English. We group part-of-speech tags into \textit{Nouns}, \textit{Adjectives}, \textit{Verbs}, and an \textit{Other} category that encompasses tags such as particles, etc. We observe that the same words used for name entities in Arabic appear at high frequencies as adjectives and verbs rather than nouns. Arabic entities also have a higher Information Entropy $\mathcal{H}$\footnote{Information Entropy in bits for $N$ POS tags: \\ $\mathcal{H} = -\sum_{i=1}^N p(\mathrm{tag}_i) \log_2 p(\mathrm{tag}_i)$} \cite{ramscar2019source}, measured at 0.799 bits for named compared with uncased English entities at 0.267 bits. It is important to note that English employs casing for name entities, facilitating a clear distinction between a name ``\textit{Mark}" and the verb ``\textit{mark}". In standard cased English text, occurrences of entities as adjectives or verbs are minimal, with entities predominantly appearing as nouns ($\mathcal{H}$ = 0.104 bits). However, Arabic does not have casing conventions, resulting in greater variability in the grammatical roles of named entities (used for descriptions as adjectives or actions as verbs), which can cause challenges for LMs in distinguishing between word senses as entities or non-entities. Conversely, named entities in English tend to adhere more consistently to noun roles.

\begin{figure*}[t!]
    \centering
    \includegraphics[width=\linewidth]{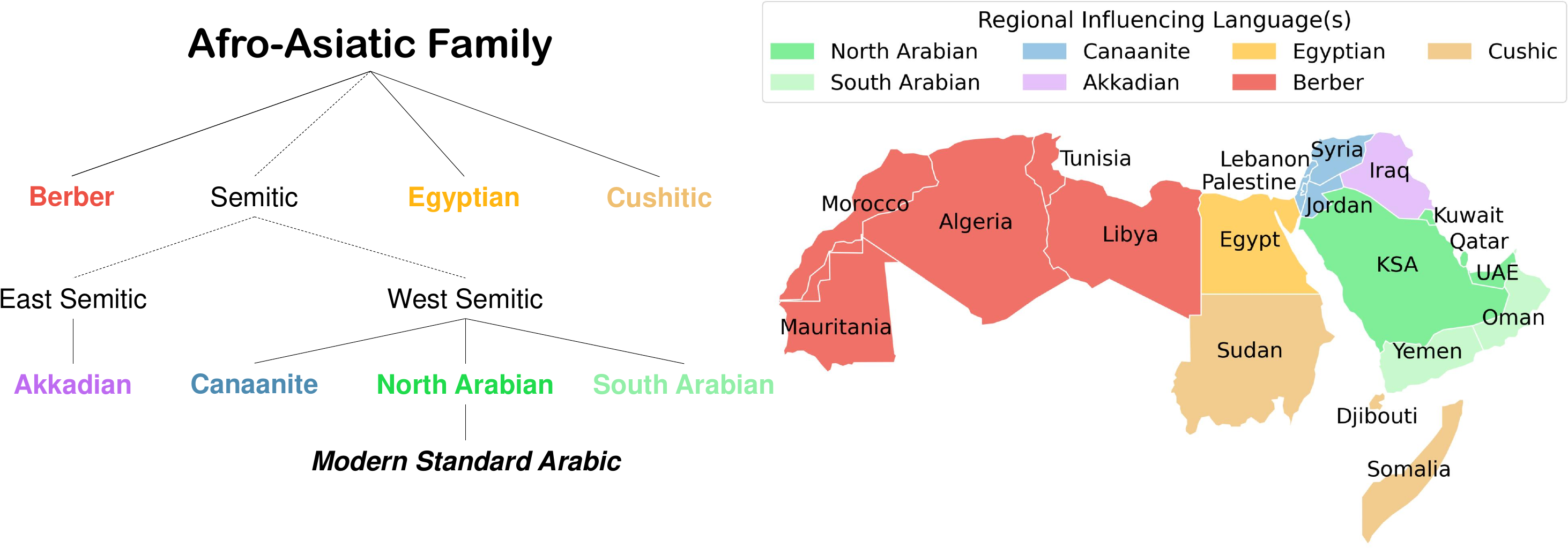}
    \caption{Map of the regional influencing languages on location names in Arab countries, and their standard classification in the Afro-Asiatic family according to \citet{versteegh2014arabic}. The \textit{Comorian} language of the Comoros Island (not shown here on the map) is outside of the Afro-Asatic family and belongs to the Niger-Congo family. }
    \label{fig:language-tree}
\end{figure*}

\subsection{Regional Influences on Arabic Entities}
\label{appendix:regional-influences}

While many named entities in the Arabic language are polysemous words, there are also words used for named entities that do not serve other functional purposes. These words are often Arabized forms of regional linguistic influences from different parts of the Arab world. A product of historical heritage, the countries of today's Arab world present a rich cultural mosaic. Following a sociological process of Arabization in North Africa and West Asia \cite{reynolds2015cambridge}, the spread of the Arabic language in those areas and its interactions with regional languages led to the development of contemporary Arabic dialects. For example, the dialects in countries of the Levant region (i.e, Lebanon, Syria, Palestine, Jordan) have been shaped by influences from other Semitic languages historically spoken in those areas such as Aramaic, Hebrew, and Canaanite languages \cite{gragg2019semitic}. The names of many of today's cities, villages, and towns in those countries originate from the regional influencing languages, which predate the spread of Arabic. For instance, the Arabic naming of the Lebanese capital, Beirut (in Arabic: \setcode{utf-8}`\<بيروت>'), is a transliterated derivation of its Phoenician name  ``bī'rōt". Such entities would only appear in the Arabic in the sense of a location and do not have any other lexical uses.

There are also regional linguistic influences on location names in the Arabian peninsula itself. Different Southern Arabian Languages were spoken in the southern part of the peninsula (i.e., present-day Yemen and Oman), while multiple Old North Arabian dialects were  spoken in the central and northern parts of the peninsula (i.e., present-day Saudi Arabia). Contemporary Arabic (i.e., Modern Standard Arabic) is a continuum from Classical Arabic, the language of the Quraan, and early Islamic literature, which was the dialect of the Quraysh tribe in central Arabia. Figure~\ref{fig:language-tree} presents a visualization of the regional linguistic influences on each Arab country and their classification within the Afro-Asiatic language family \cite{versteegh2014arabic}.

\section{Analyses: Additional Results}

\subsection{Entity Frequency in Pre-training Data}
\label{appendix:entity-frequency-results}

\begin{figure}
    \centering
    \includegraphics[width=\linewidth]{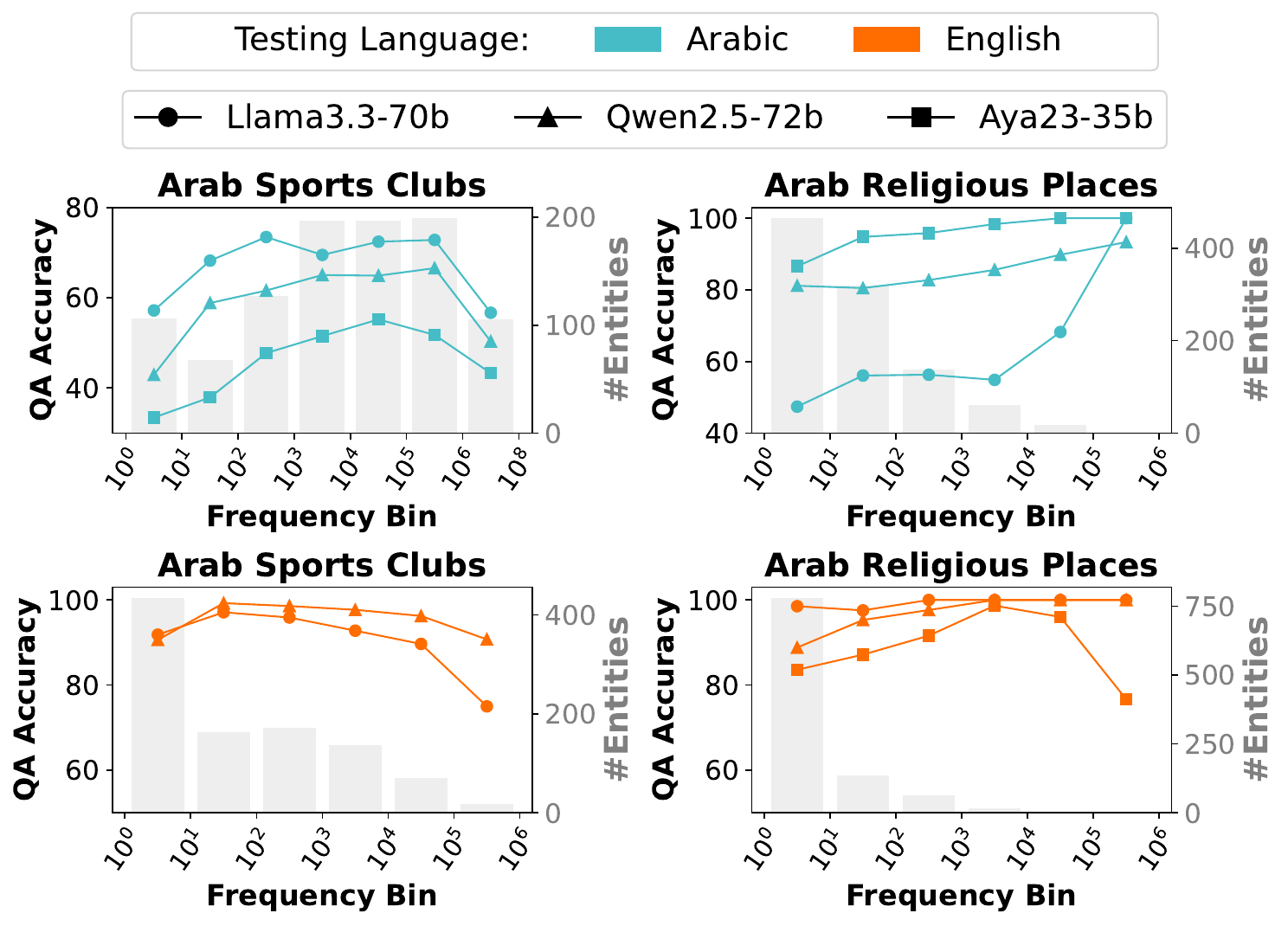}
    \caption{Average QA Accuracy ($\uparrow$) of LLMs when tested in Arabic and English on sports clubs and religious places of worship associated with Arab culture, stratified by their occurrence counts in the mC4 corpus (grouped into log10-spaced bins).}
    \label{fig:frequency-qa-sports-religious}
\end{figure}

\begin{figure*}[t]
    \centering
    \includegraphics[width=0.9\linewidth]{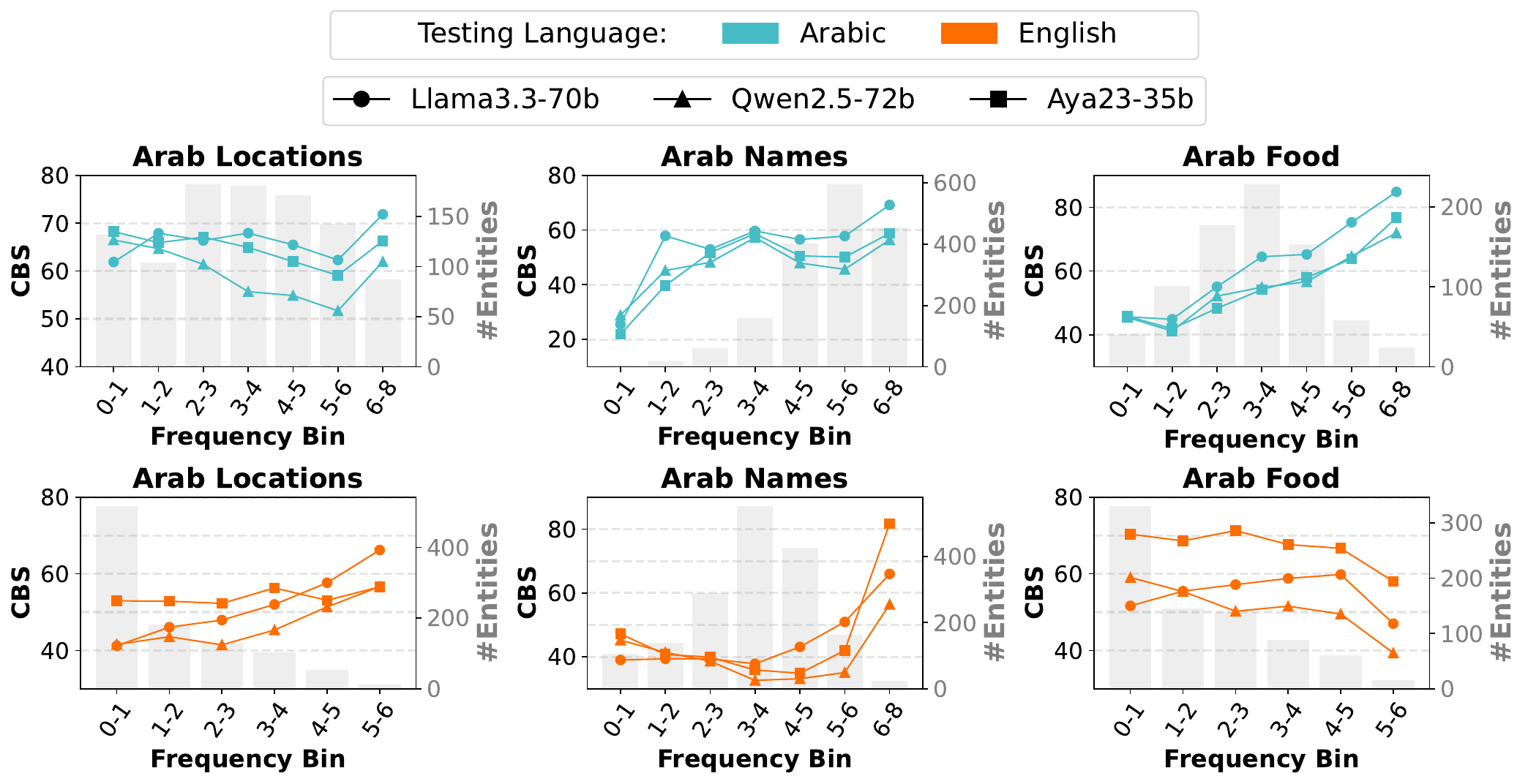}
    \caption{Average CBS ($\downarrow$) of LLMs at text-infilling when tested in Arabic and English on locations, names, and food entities associated with Arab culture, stratified by their occurrence counts in the mC4 corpus (grouped into log10-spaced bins}
    \label{fig:cbs-frequency-analysis}
\end{figure*}

\paragraph{Extractive QA.} Figure~\ref{fig:frequency-qa-sports-religious} shows the performance of LMs on extractive QA of sports clubs and religious places of worship stratified by their occurrence in Arabic and English pre-training data. We observe similar trends to our results in \S\ref{subsec:corpus-frequencies}.

\paragraph{Text Infilling.} Figure~\ref{fig:cbs-frequency-analysis} shows the average CBS at the text-infilling task (\S\ref{subsec:text-infilling-task}) on entities stratified by their occurrence counts in pre-training. In this setup, we test each Arab entity against 30 randomly sampled Western entities across 5 randomly sampled culturally-contextualized contexts from CAMeL-2. Similar to our observations in \S\ref{subsec:corpus-frequencies}, we find that models in Arabic struggle on high-frequency entities, where CBS is the highest for all entity types. This indicates that entities that appear at very high frequencies will be assigned lower likelihood than Western entities given contexts grounded in Arab culture. We also observe a struggle on highly frequent entities when the model is tested in English on Locations and Names, which are rare cases where translations of Arabic entities exhibit polysemy. For example, this happens with the feminine name  \setcode{utf-8}``\<آسيا>'' that is written as "\textit{Asia}" in English, matching the name of the continent.

\subsection{The Impact of Word Polysemy}
\label{appendix:polysemy-results}

\begin{figure*}[t]
    \centering
        \includegraphics[width=\linewidth]{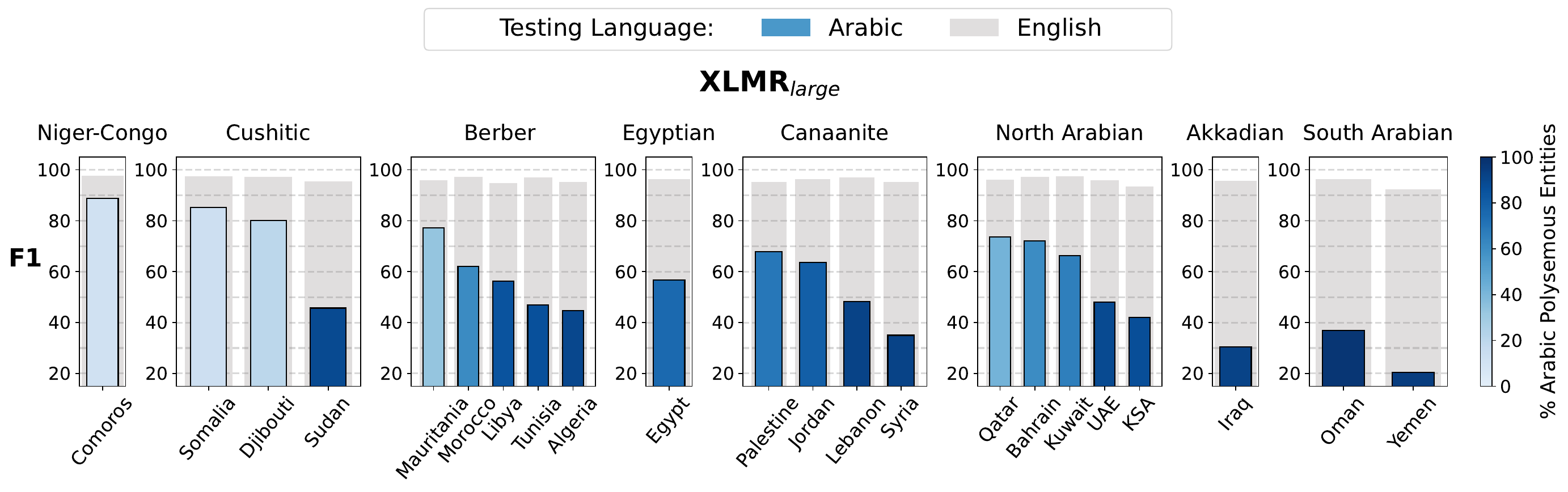}
        \includegraphics[width=\linewidth]{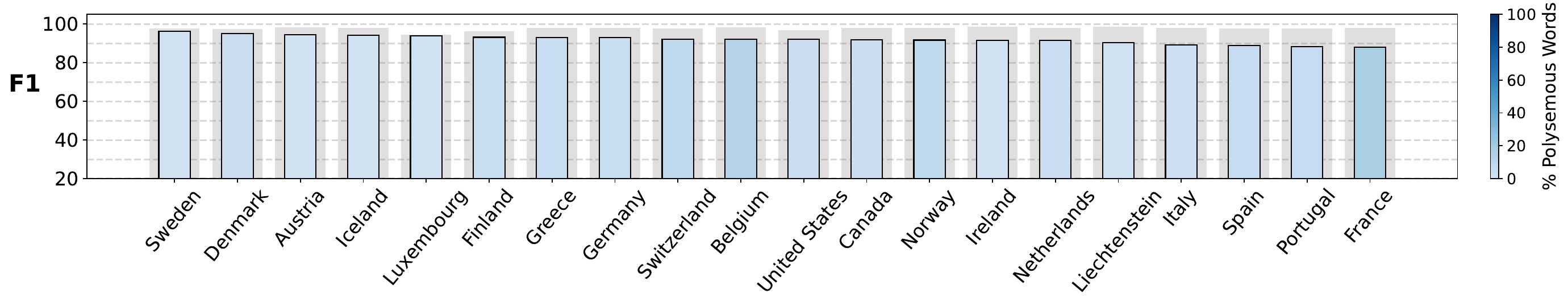}

    \caption{Average NER F1 of XLMR$_{\mathrm{large}}$ on the top-100 most frequent location entities in mC4 for each Arab country (top) and Western country (bottom) in CAMeL-2.}
    \label{fig:xlmr-top100-locations}
\end{figure*}

\begin{figure*}[t]
    \centering
        \includegraphics[width=\linewidth]{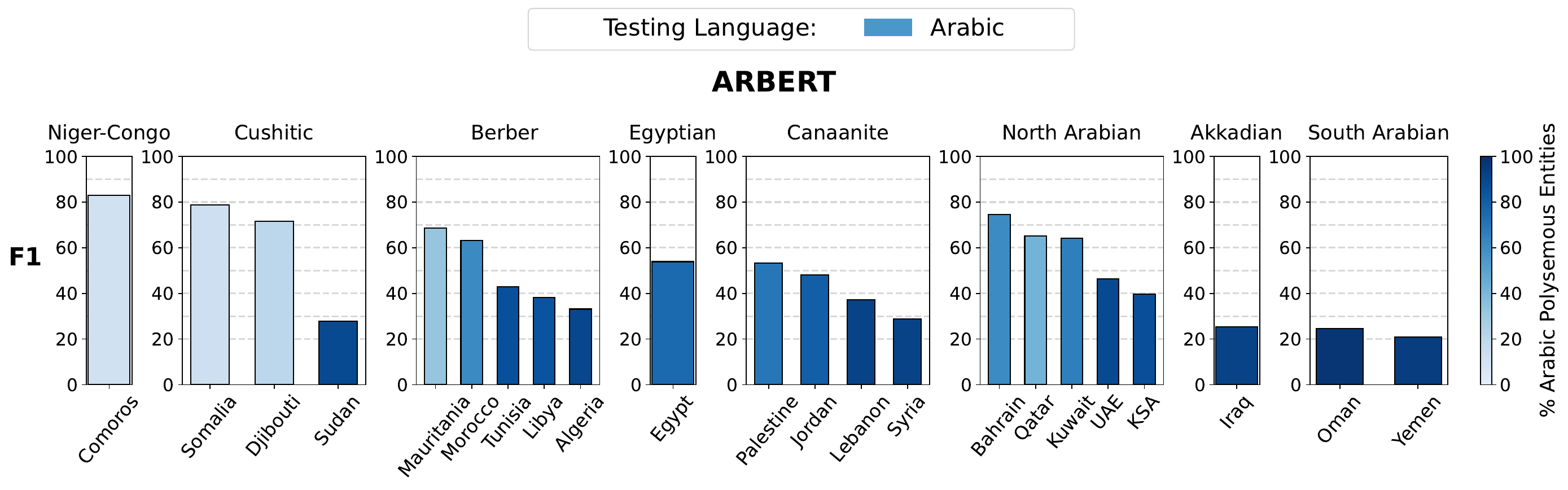}
        \includegraphics[width=\linewidth]{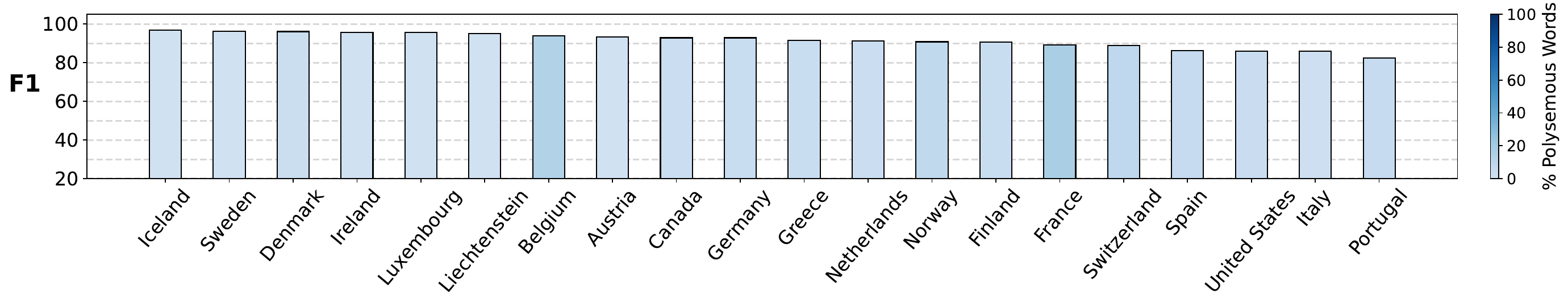}
    \caption{Average NER F1 of ARBERT on the top-100 most frequent location entities in mC4 for each Arab country (top) and Western country (bottom) in CAMeL-2.}
    \label{fig:arbert-top100-locations}
\end{figure*}

\paragraph{NER.} Figure~\ref{fig:xlmr-top100-locations} and Figure~\ref{fig:arbert-top100-locations} show the results XLMR$_{\mathrm{large}}$ and ARBERT on NER of the top-100 Arab and Western locations as a function of the percentage of entities that match Arabic polysemous words.  We find the same trend observed with Llama3.3-70b on extracted QA, where performance becomes poor on entities that exhibit word polysemy in Arabic (\S\ref{subsec:entity-polysemy-analysis}).

\subsection{The Impact of Tokenization}
\label{appendix:tokenization-results}

\begin{figure*}[t]
    \centering
    \includegraphics[width=\linewidth]{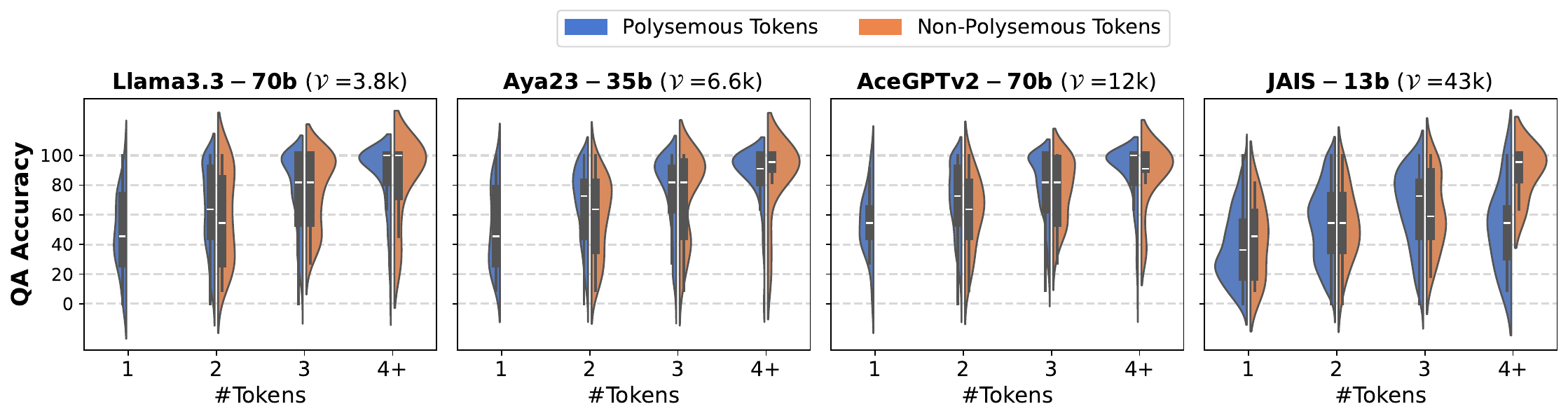}
    \includegraphics[width=\linewidth]{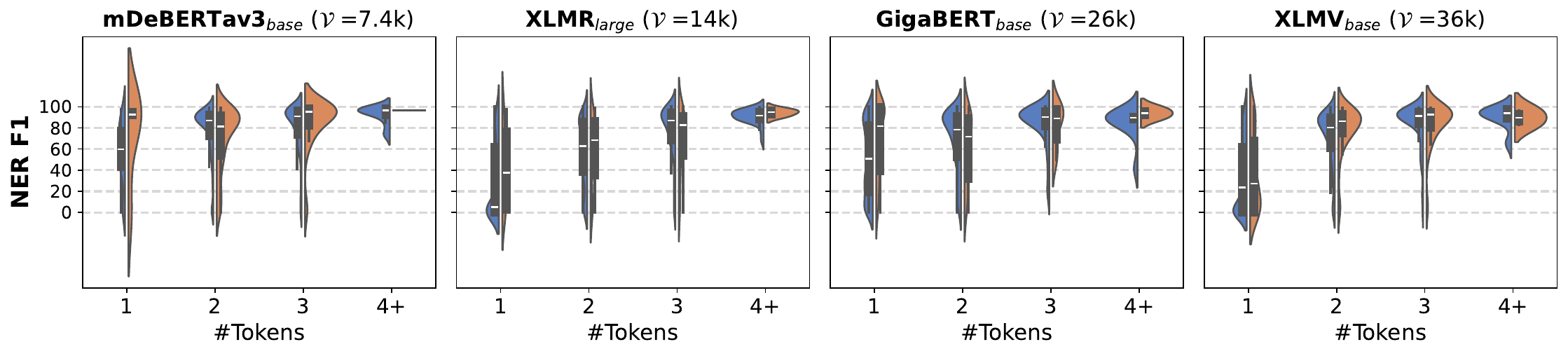}
    \includegraphics[width=\linewidth]{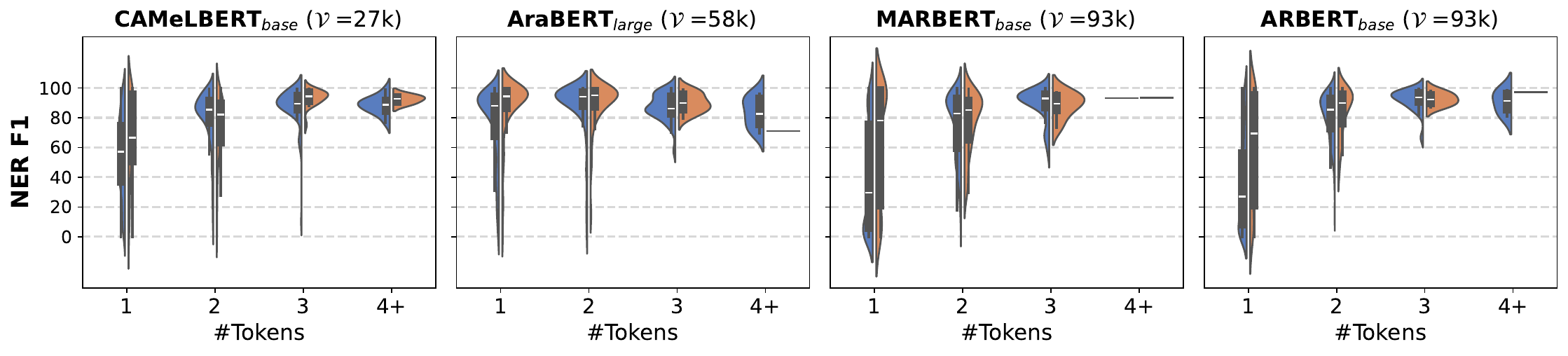}
    \caption{Performance distribution of several LMs on Arab location entities as a function of how many tokens they get tokenized into. Entities are separated based on whether tokens correspond to Arabic polysemous words. $\mathcal{V}$ represents the number of Arabic tokens in each LM's vocabulary.}
    \label{fig:tokenization-all-models}
\end{figure*}

We report the performance of a variety of LMs on location entities as a function of how many tokens they are fragmented into in Figure~\ref{fig:tokenization-all-models}. Similar to our observations in \S\ref{subsec:tokenization-analysis}, we find that performance improves as entities are tokenized into more than a single token, and that LMs struggle with one-token entities that exhibit word polysemy. We also see that this issue gets worse as the number of Arabic tokens in an LM's vocabulary gets larger.

\end{document}